\RequirePackage{fix-cm}
\documentclass[smallextended]{svjour3}
\smartqed  % flush right qed marks, e.g. at end of proof
\journalname{Cognitive Computation}

\usepackage[immediate]{silence}
\usepackage{hyperref}
\usepackage{natbib}\let\cite\citep
\usepackage{gb4e}
\noautomath
\usepackage{times}
\usepackage{latexsym}
\usepackage{amsmath}

\usepackage[capitalize]{cleveref}

\usepackage{amssymb}
\usepackage{multirow,multicol}
\usepackage{graphicx}
\graphicspath{{figs/}}

\usepackage{array}
\usepackage{soul,color}
    
    \usepackage{booktabs}

\usepackage[colorinlistoftodos]{todonotes}
\usepackage{enumitem}
\usepackage{subcaption}
\usepackage{tipa}
\usepackage{stackengine}
\usepackage{rotating}
\usepackage{tabularx}
\newcommand{\PreserveBackslash}[1]{\let\temp=\\#1\let\\=\temp}
\newcolumntype{C}[1]{>{\PreserveBackslash\centering}p{#1}}
\newcolumntype{R}[1]{>{\PreserveBackslash\raggedleft}p{#1}}
\newcolumntype{L}[1]{>{\PreserveBackslash\raggedright}p{#1}}

\definecolor{gelbukh}{rgb}{.9,1,.9} 
\newcommand\emo[1]{\textsc{#1}}
\newcommand\code[1]{\texttt{#1}}
\newcommand\ToDeleteIfNeedSpace{}

\newcommand\RECCON{recognizing emotion cause in conversations} % \emph{} removed

\newcommand\RECCONDA{RECCON}
\newcommand\RECCONDADD{RECCON-DD}
\newcommand\RECCONDAIE{RECCON-IE}
\newcommand\DailyDialog{Daily\-Dialog}
\newcommand\0{\hphantom{0}}

\begin{document}

% \title{RECCON: A Dataset for Recognizing Emotion Cause in CONversations}
\title{Recognizing Emotion Cause in Conversations%
\thanks{%
S. Poria, N. Majumder, D. Ghosal, R. Bhardwaj, S. Yu Bai Jian, and P. Hong have received support from the  A*STAR under its RIE 2020 Advanced Manufacturing and Engineering programmatic grant, Award No.~A19E2b0098.
A. Gelbukh has received support from the Mexican Government through the grant A1-S-47854 of the CONACYT, Mexico, and grants 20211784, 20211884, and 20211178 of the Secretar\'{\i}a de Investigaci\'on y Posgrado of the Instituto Polit\'ecnico Nacional, Mexico.
}}
% RM: I like a little better this title: is stronger, as it indicates the introduction of the task

\author{%
     Soujanya Poria
\and Navonil Majumder
\and Devamanyu Hazarika
\and Deepanway Ghosal
\and Rishabh Bhardwaj
\and Samson Yu Bai Jian
\and Pengfei Hong
\and Romila Ghosh
\and Abhinaba Roy
\and Niyati Chhaya
\and Alexander Gelbukh\thanks{Corresponding author: Alexander Gelbukh} 
\and Rada Mihalcea}

% \newcommand\aand{\unskip,\hspace{0.09ex}} % 0.31
% \authorrunning{Poria \aand Majumder \aand Hazarika \aand Ghosal \aand Bhardwaj \aand Jian \aand Hong \aand Ghosh \aand Roy \aand Chhaya \aand Gelbukh \aand Mihalcea}

\institute{%
S. Poria, N. Majumder, D. Ghosal, R. Bhardwaj, S. Yu Bai Jian, and P. Hong \at
Singapore University of Technology and Design, Singapore \\
% \email{\{sporia@, navonil\_majumder@, deepanway\_ghosal@mymail., rishabh\_bhardwaj@mymail., samson\_yu@, pengfei\_hong@mymail.\}sutd.edu.sg}
\email{sporia@sutd.edu.sg, navonil\_majumder@sutd.edu.sg, deepanway\_ghosal@mymail.sutd.edu.sg, rishabh\_bhardwaj@mymail.sutd.edu.sg, samson\_yu@sutd.edu.sg, pengfei\_hong@mymail.sutd.edu.sg}
\and
D. Hazarika \at
National University of Singapore, Singapore \\
\email{hazarika@comp.nus.edu.sg}
\and
R. Ghosh \at
Independent researcher, India \\
\email{romila.ghosh93@gmail.com}
\and
A. Roy \at
Nanyang Technological University, Singapore \\
\email{abhinaba.roy@ntu.edu.sg}
\and
N. Chhaya \at
Adobe Research, India \\
\email{nchhaya@adobe.com}
\and
A. Gelbukh (corresponding author) \at
CIC, Instituto Polit\'ecnico Nacional, Mexico \\
\email{gelbukh@gelbukh.com}
\and
R. Mihalcea \at
University of Michigan, USA \\
\email{mihalcea@umich.edu}
}

%%%%%%%%%%%%%%%%%%%%%%%%%%%%%%%%%%%%%%%%%%%%%%%%%%
%%%%%%%%%%%%%%%%%%%%%%%%%%%%%%%%%%%%%%%%%%%%%%%%%%
%%%%%%%%%%%%%%%%%%%%%%%%%%%%%%%%%%%%%%%%%%%%%%%%%%
%\Thanks{~Equal contribution. Randomly ordered.}~%

%%%%%%%%%%%%%%%%%%%%%%%%%%%%%%%%%%%%%%%%%%%%%%%%%%
%%%%%%%%%%%%%%%%%%%%%%%%%%%%%%%%%%%%%%%%%%%%%%%%%%
%%%%%%%%%%%%%%%%%%%%%%%%%%%%%%%%%%%%%%%%%%%%%%%%%%
% \corrauthor[1]{First Author}{f.author@email.com}

%%%%%%%%%%%%%%%%%%%%%%%%%%%%%%%%%%%%%%%%%%%%%

%\author{Soujanya Poria$^1$, 
%Navonil Majumder$^1$, Devamanyu Hazarika$^2$%
%%\Thanks{~Equal contribution. Randomly ordered.}~%
%,\\
%Deepanway Ghosal$^1$\footnotemark[1]~,
%% \textbf{
%Rishabh Bhardwaj$^1$, Samson Yu Bai Jian$^1$,
%\\
%Pengfei Hong$^1$, Romila Ghosh$^4$,
%% }%
%% \\
%% \textbf{
%Abhinaba Roy$^7$,\\
%Niyati Chhaya$^6$, Alexander Gelbukh$^3$, Rada Mihalcea$^5$
%% }%
%\\[1ex]
%$^1$~Singapore University of Technology and Design, Singapore\\
%$^2$~National University of Singapore, Singapore  \\
%$^3$~CIC, Instituto Politécnico Nacional, Mexico  \\
%$^4$~Independent researcher, India\\
%$^5$~University of Michigan, USA\\
%$^6$~Adobe Research, India\\
%$^7$~Nanyang Technological University, Singapore
%}

\maketitle

% \begin{abstract}
%\?{The word ``recognizing'' seems to promise more than we deliver: logically inferring vs. labeling a passage. In RTE it's about logical inference.}
% Recognizing the cause behind emotions in text is a fundamental yet under-explored area of research in NLP. Advances in this area hold the potential to improve interpretability and performance in affect-based models. Identifying emotion causes at the utterance level in conversations is particularly challenging due to the intermingling dynamics among the interlocutors. To this end, we introduce the task of \RECCON{} with an accompanying dataset named \RECCONDA. Furthermore, we define different cause types based on the source of the causes, and establish strong Transformer-based baselines to address two different sub-tasks on RECCON: \textit{$(i)$}~causal span extraction and \textit{$(ii)$}~causal emotion entailment. The dataset is available at \url{https://github.com/declare-lab/RECCON}.
% \end{abstract}

\begin{abstract}
We address the problem of recognizing emotion cause in conversations, define two novel sub-tasks of this problem, and provide a corresponding dialogue-level dataset, along with strong Transformer-based baselines.
The dataset is available at \url{https://github.com/declare-lab/RECCON}.

\paragraph{Introduction}
Recognizing the cause behind emotions in text is a fundamental yet under-explored area of research in NLP. Advances in this area hold the potential to improve interpretability and performance in affect-based models. Identifying emotion causes at the utterance level in conversations is particularly challenging due to the intermingling dynamics among the interlocutors. 

\paragraph{Method}
We introduce the task of Recognizing Emotion Cause in CONversations with an accompanying dataset named \RECCONDA, containing over 1,000 dialogues and 10,000 utterance cause-effect pairs. Furthermore, we define different cause types based on the source of the causes, and establish strong Transformer-based baselines to address two different sub-tasks on this dataset: causal span extraction and causal emotion entailment.

\paragraph{Result}
Our Transformer-based baselines, which leverage contextual pre-trained embeddings, such as RoBERTa, 
% outshine 
outperform
the state-of-the-art emotion cause extraction approaches
% ---ECPE-MLL~\cite{DBLP:conf/emnlp/DingXY20} and ECPE-2D~\cite{DBLP:conf/acl/DingXY20} on the RECCON 
on our dataset.

\paragraph{Conclusion}
We introduce a new task highly relevant for (explainable) emotion-aware artificial intelligence: recognizing emotion cause in conversations, provide a new highly challenging publicly available dialogue-level dataset for this task, and give strong baseline results on this dataset.
\end{abstract}

\section{Introduction}

Emotions are intrinsic to humans; consequently,
emotion understanding is a key part of human-like artificial
intelligence (AI). Language is
often indicative of one's emotions. Hence, emotion recognition has attracted much attention 
% been enjoying popularity 
in the field of natural language processing (NLP)~\citep{kratzwald2018decision, colneric2018emotion} due to its wide range of applications in
opinion mining, recommender systems, healthcare, and other areas.

In particular, emotions are an integral part of human cognition; thus understanding human emotions and reasoning about them is one the key issues in computational modeling of human cognitive processes~\citep{Izard1992}. Among different settings where human emotions play important cognitive role is human-human and human-computer conversations. Similarly, among different issues in automatic reasoning about human emotions is identifying the causal root of the expressed emotions in the discourse of such a conversation.
During a dialog, cognitive and affective processes can be triggered by non-verbal external events or sensory input. % In the process of a conversation, certain sensory or other external events can directly initiate cognition and affect. These inputs can often be non-verbal cues. 
Sometimes such affective processes can happen even before the corresponding cognitive processing by the person---a phenomenon called \emph{affective primacy}~\citep{Zajonc80feelingand}. % Affective reactions to these sensory inputs can occur with or without any complex cognitive modeling. When the stimulus is sudden and unexpected, the affective reaction can occur before evaluating and appraising the situation through cognitive modeling. This is called \emph{Affective Primacy}~\citep{Zajonc80feelingand}. For example, our immediate reaction when we encounter an unknown creature in the jungle without evaluating whether it is safe or dangerous. 
On the other hand, complex cognitive processing, which would lead to updating the computational model's speaker state, can also happen before or after the affective reaction of the participant of the conversation. % Sensory inputs can also trigger cognitive modeling for subsequent evaluation of the situation and thus can update the latent speaker-state. 

Substantial progress has been made in the detection and classification of emotions, expressed in text or videos, according to emotion taxonomies~\cite{ekman1993facial,plutchik}. However, further reasoning about emotions, such as  
understanding the cause of an emotion expressed by a speaker,
has been less explored so far.
For example, 
% consider 
understanding
the following review of a smartphone, ``\textit{I hate the touchscreen as it freezes after 2-3 touches}'',
% .
% (emotion: \emo{disgust}).
% \?{this is not about a conversation? Soujanya: we first start with monologues and then shift to dialogues}
% Understanding this text 
implies not only detecting the expressed negative emotion, specifically \emo{disgust}, but also spotting its cause~\citep{liu2012sentiment}---in this case, ``\textit{it freezes after 2-3 touches}.''

Of a wide spectrum of emotion-reasoning tasks~\cite{ellsworth2003appraisal},
in this work we focus on identifying the causes (also called antecedents, triggers, or stimuli) of emotions expressed specifically in conversations. In particular, we look for events, situations, opinions, or experiences in the conversational context that are primarily responsible for an elicited emotion in the target utterance. Apart from event mentions, the cause could also be a speaker's counterpart reacting towards an event cared for by the speaker (inter-personal emotional influence). 

We introduce the task of \textbf{\underline{r}ecognizing \underline{e}motion \underline{c}ause in \underline{con}\-ver\-sa\-tions} (RECCON), which refers to the extraction of such stimuli behind an emotion in a conversational utterance. The cause could be present in the same or contextual utterances.
%AG (conversational history). 
We formally define this task in~\Cref{sec:annot}.

\begin{figure}[t] 
    \centering 
    \includegraphics[width=60ex]{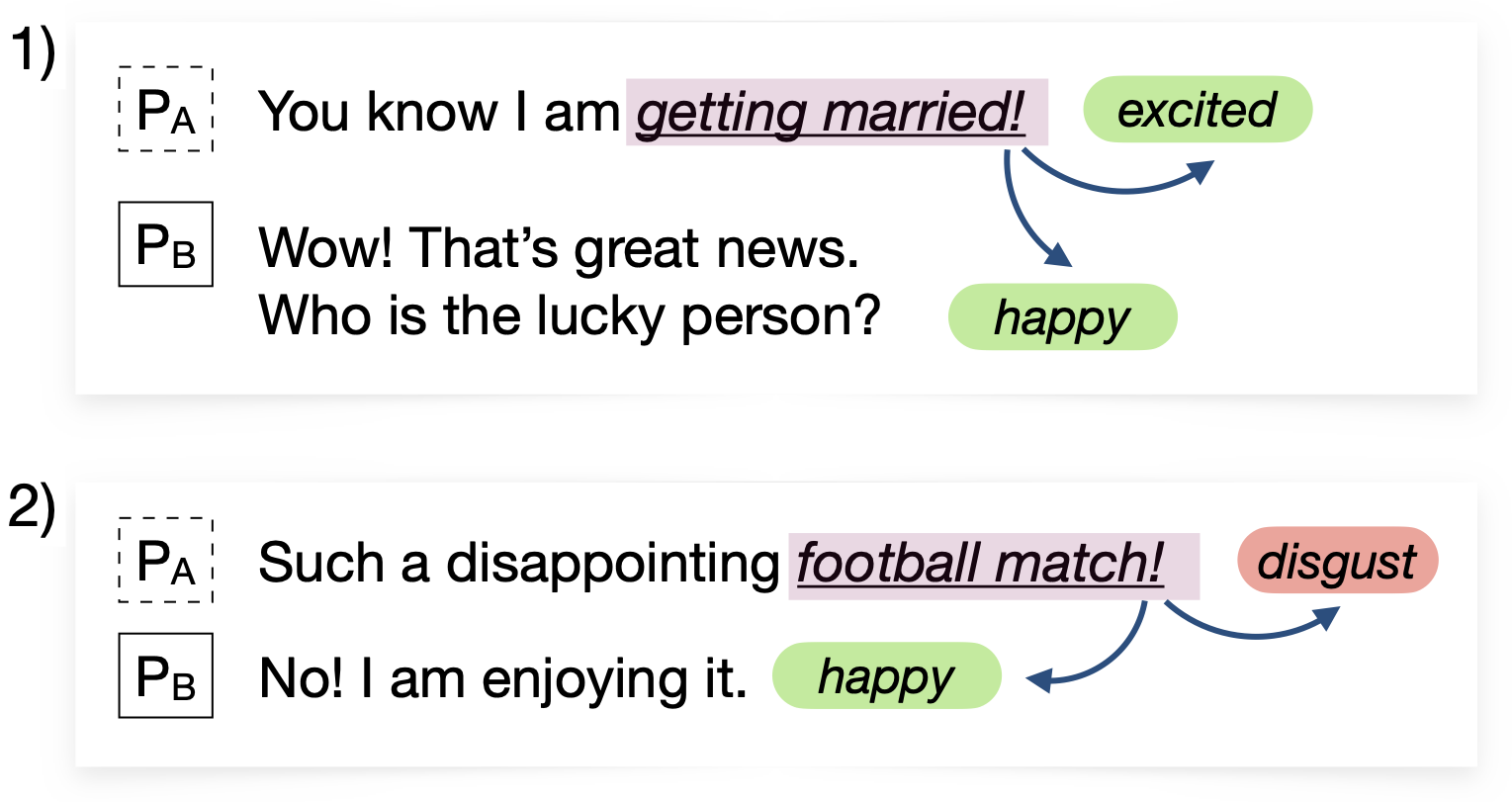}
    \caption{Emotion causes in conversations.}
    \label{fig:examples}
\end{figure}

In~\cref{fig:examples} we exemplify this task. In the first example, we want to know the cause of person B's ($P_B$) emotion (\emo{happy}). It can be seen that $P_A$ is happy due to the event ``\textit{getting married}'' and similarly $P_B$ also reacts positively to this event. Here, we can infer that $P_B$'s emotion is caused either by the reference of the first utterance to the event of getting married or by the fact that $P_A$ is happy about getting married---both of which can be considered as stimulus for $P_B$'s emotion. In the second example, the cause of $P_A$'s emotion is the event ``\textit{football match}" and a negative emotion \emo{disgust} indicates that $P_A$ is unsatisfied 
% experience of 
with
the match. In contrast, $P_B$ 
% takes pleasure of 
likes
the match---sharing the same cause with $P_A$---with \emo{happiness} emotion. These examples demonstrate the challenging problem of recognizing emotion causes in conversations, which to the best of our knowledge, is one of the first attempts in this area of research.

We can summarize our contributions as follows:

% \?{I don't understand how this para is relevant: it is about }\cref{def:emo-reason}\?{ which is left to future work anyway.}
% It is hard to define a taxonomy or tagset for the causal reasoning of emotions. At present, there is no available dataset that would contain such rich annotations in conversation level. Building such a dataset can enable future dialogue systems to frame meaningful argumentation logic and discourse structure, taking one step closer to human-like conversation.

\begin{enumerate}%[itemsep=0ex,leftmargin=*]

    \item We introduce a new task, \textbf{\RECCON{}}, and dive into many unique characteristics of this task that is peculiar to conversations. 
    %
    % of \textbf{R}ecognizing \textbf{E}motion \textbf{C}ause in \textbf{CON}versations (RECCON)\footnote{RECCON is pronounced as /\textquotesingle r$\varepsilon$ k(\textipa{@})n/.}, 
    %\?{can be RECiC /rekik/ or REC-C /reksi/}
    In particular, we define the relevant types of emotion causes (\cref{sec:types}). 
     
     \item 
     Further, 
     we describe a new annotated dataset for this task, \RECCONDA{}\footnote{pronounced as \textit{reckon}.}, including both acted and real-world dyadic conversations (\cref{sec:dataset}). To the best of our knowledge, there is no other dataset for the task of emotion cause recognition \emph{in conversations}.
     
     \item 
     % Further, 
     Finally, 
     we introduce two challenging sub-tasks that demand complex reasoning, and 
     % provide the corresponding baselines (\cref{sec:experiments}).
     % \item we 
     setup strong baselines to solve the sub-tasks (\cref{sec:experiments}). These baselines surpass the performance of several newly introduced complex neural approaches, e.g., ECPE-MLL~\cite{DBLP:conf/emnlp/DingXY20}, RankCP~\cite{wei-etal-2020-effective},  and ECPE-2D~\cite{DBLP:conf/acl/DingXY20}.
\end{enumerate}

\section{Related Work} \label{sec:related_works}

Initial works on emotion analysis were applied to the opinion mining task, exploring different aspects of affect beyond polarity prediction, such as identifying the opinion~/~emotion feeler (holder, source)~\cite{das-bandyopadhyay-2010-finding,DBLP:conf/naacl/ChoiCRP05}.
More recently, sentiment analysis research has been used in a wider context of natural language understanding and reasoning~\citep{OntoSenticNet2}.
New methods for multi-label emotion classification are developed~\cite{Iqra} and new corpora for emotion detection are compiled for languages other than English~\cite{LiSSS}.

% However, 
The task of emotion cause extraction was 
% later 
studied initially by~\citet{lee-etal-2010-text}. 
% Such initial 
The early works 
% involved extracting cause events in a 
used 
% a rule-driven manner
rule-based
approaches~\cite{chen-etal-2010-emotion}. \citet{gui2016event} constructed an emotion cause extraction dataset by identifying events that trigger emotions. They used news articles as their source for the corpus to avoid the latent emotions and implicit emotion causes associated with the informal text, thus reducing reasoning complexity for the annotators while extracting emotion causes. Other notable works on emotion cause extraction (ECE) are \citep{DBLP:conf/cicling/GhaziIS15} and \citep{gao2017overview}.

% \begin{itemize}
%   % \item  \citet{gui2016event} introduced the Chinese dataset.
%     % \item \cite{DBLP:conf/emnlp/FanYDGBYXM19} Knowledge-regularized ECE. Uses discourse information and knowledge from lexicons as inductive biases (for regularizing).
%     \item \cite{DBLP:conf/cicling/GhaziIS15}
%     \item \cite{gao2017overview} proposed a shared task for ECA on both Chinese and English text from novels.
%     % \item \cite{DBLP:conf/emnlp/SongBGLZWSLZ19} annotate previous utterances sentiment triggers. However, some critical distinctions are - 1) They do this for customer-service calls which is highly topical 2) we do not fine-grained annotation as they just mark the utterances (Our task 1). \#TODO: Check if they do multi-utterance annotations.\hl{Soujanya: I read this paper and it is similar to ERC with just one difference - it classifies the whole dialogue.}
% \end{itemize}

% \citet{DBLP:conf/emnlp/FanYDGBYXM19} attempted to solve ECE by proposing a Knowledge-regularized model by leveraging discourse information and knowledge from lexicons as inductive biases to regularize the network. 

As a modification of the ECE task, \citet{DBLP:conf/acl/XiaD19} proposed emotion-cause pair extraction (ECPE) that jointly identifies both emotions and their corresponding causes~\cite{DBLP:conf/emnlp/ChenHCL18}. Further, \citet{chen-etal-2020-conditional} recently proposed the conditional emotion cause pair (ECP) identification task, where they highlighted the causal relationship to be valid only in particular contexts. We incorporate this property in our dataset construction, as we annotate multiple spans in the conversational history that \textit{sufficiently} indicate the cause. Similar to~\citet{chen-etal-2020-conditional}, we also provide negative examples of context that does not contain the causal span. 

% \citet{li2017reflections} discussed two possible reasons that give rise to opinions: firstly, an opinion holder might have an emotional bias towards the entity or topic in question; secondly, the emotion could be borne out of mental (dis)satisfaction towards a goal achievement. 
Our work is a natural extension of those works. We propose a new dataset on dyadic conversations, which is more difficult to annotate. Additionally, the associated task of recognizing emotion cause in conversations poses a greater hitch to solve due to numerous challenges. For example, \textit{$(i)$}~expressed emotions are not always explicit in the conversations; \textit{($ii)$}~conversations can be very informal where the phrase connecting emotion with its cause can often be implicit and thus needs to be inferred; \textit{$(iii)$}~the stimuli of the elicited emotions can be located far from the target utterance in the conversation history, so that detecting it requires complex reasoning and co-reference, often using commonsense.

\section{Definition of the Task}
\label{sec:terminology}

We distinguish between emotion \textbf{evidence} and emotion \textbf{cause}:
\begin{itemize}[itemsep=0ex,leftmargin=*]
    \item \textit{Emotion evidence} is a part of the text that indicates the presence of an emotion in the speaker's emotional state. It acts in the real world between the text and the reader. 
    % (or the system). 
    Identifying and interpreting the emotion evidence is the underlying process of the well-known emotion detection task. 
    
    \item \textit{Emotion cause} is a part of the text expressing the reason for the speaker to feel the emotion given by the emotion evidence. It acts in the 
    % virtual world described by the text 
    described world
    between the (described) circumstances and the (described) speaker's emotional state. Identifying the emotion cause constitutes the task we consider in this~paper.
\end{itemize}
For instance, in \cref{fig:examples}, $P_B$'s turn contains evidence of $P_B$'s emotion, while $P_A$'s turn contains its cause.
The same text span can be both emotion evidence and cause, but generally this is not the~case.

Defining the notion of emotion cause is, in a way, the main goal of this paper. However, short of a formal definition, we will explain this notion on numerous examples and, in computational terms, via the labeled dataset. 
%AG Note that a text part can be both emotion evidence and cause.

We use the following terminology throughout the paper. 
The \textbf{target utterance} $U_t$ is the $t^{th}$ utterance of a conversation, whose emotion label $E_t$ is known and whose emotion cause we want to identify.
The \textbf{conversational history} $H(U)$ of the utterance $U$ is the set of all utterances from the beginning of the conversation till the utterance $U$, including $U$.
A \textbf{causal span} for an utterance $U$ is a maximal sub-string, of an utterance from $H(U)$, that is a part of $U$'s emotion cause; we will denote the set of the causal spans for an utterance $U$ by $CS(U)$.
A \textbf{causal utterance} is an utterance containing a causal span; we denote the set of all causal utterances for $U$ by $C(U)\subseteq H(U)$.
% = \{U^t_c\ \mid c \leq t\}$ and $C_{U_t} \subseteq H_{U_t}$.
% \textbf{Non-causal utterance ($U^t_n$):} 
% An utterance which \textit{does not} contain a causal span for $U_t$. We represent the set of all non-causal utterances for $U_t$ as $\Tilde{C}_{U_t} = \{U^t_n\ \mid \forall U^t_n \notin C_{U_t}\}$. \\
An \textbf{\underline utterance--\underline causal \underline span (UCS) pair} is a pair $(U,S)$, where $U$ is an utterance and $S\in CS(U)$.

Thus, \textbf{recognizing emotion cause} is the task of identifying all (correct) UCS pairs in a given~text.

% $(U_t, S_c)$ refers to the pair of target utterance $U_t$ and one of its causal spans $S \in CS(U_t)$. Here, $S \in CS(U_t)$ is a sub-string of $U^t_c$. 
In the context of our training procedure, we will refer to (correct) UCS pairs as \textbf{positive 
% (valid) 
examples}, whereas pairs $(U, S)$ with $S\notin CS(U)$ are  \textbf{negative
% ~/~Invalid 
examples}.
% for our training process. 
In \cref{sec:neg}, we describe the sampling strategies for negative examples.

\section{Building the \RECCONDA{} dataset}\label{sec:dataset}

\subsection{Emotional Dialogue Sources}
We consider two popular conversation datasets \textbf{IEMOCAP}~\cite{iemocap} and \textbf{\DailyDialog{}}~\cite{li2017DailyDialog}, both equipped with utterance-level emotion labels:

% \begin{table}[ht!]
% \centering
% \small
% \begin{tabular}{l|cc|}
% \toprule
% \textbf{Emotion}    & \textbf{\RECCONDADD} & \textbf{\RECCONDAIE{}} \\
%             \midrule
% Anger       & 451 & 89 \\
% Fear        & 74 & - \\
% Disgust     & 140 & - \\
% Frustration &  - & 109 \\
% Happy       & 4361 & 58 \\
% Sad         & 351 & 70 \\
% Surprise    & 484 & - \\
% Excited     & - & 197 \\
% Neutral     & 5243 & 142 \\   
% \bottomrule
% \end{tabular}
% 	\caption{{Emotion label distribution in \RECCONDA.}}
% 	\label{tab:datasetdist}
% \end{table}

\begin{description}
\item[\textbf{IEMOCAP}] is a dataset of two-person conversations in English annotated with six emotion classes: \emo{anger}, \emo{excited}, \emo{frustrated}, \emo{happy}, \emo{neutral}, 
% and 
\emo{sad}. The dialogues in this dataset span across sixteen 
% unique 
conversational situations. To avoid redundancy, we handpicked only one dialogue from each of these situations. We denote the subset of our 
% \RECCONDA{} 
dataset comprising these dialogues as 
\RECCONDAIE{}.

\item[\textbf{\DailyDialog{}}] is an English-language natural human communication dataset covering various topics on our daily lives. All utterances are labeled with emotion categories: \emo{anger}, \emo{disgust}, \emo{fear}, \emo{happy}, \emo{neutral}, \emo{sad}, 
% and 
\emo{surprise}. Since the dataset is skewed 
% (over 
($83\%$ \emo{neutral} labels),
% . Due to this skewness, 
we randomly selected dialogues 
% which had 
with
at least four non-\emo{neutral} utterances. We denote the subset of \RECCONDA{} comprising these dialogues from \DailyDialog{} as \RECCONDADD. Some statistics about the annotated dataset is shown in \cref{tab:stat}.
\end{description}
Thus our RECCON dataset consists of two parts, \RECCONDAIE{} and \RECCONDADD{}. In particular, the label sets are slightly different in these two parts, as explained above.
\paragraph{Why
% Need for 
sampling from two 
% different 
datasets} 
\label{sec:dataset_diffs}
Although both IEMOCAP and \DailyDialog{} are annotated with utterance-level emotions, they differ in many aspects. First, 
% the average number of utterances per dialogue in IEMOCAP is more than $50$, 
IEMOCAP has more than $50$ utterances per dialogue on average,
whereas \DailyDialog{} has 
% a smaller average length of 
only~$8$ on average. Second, the shifts between non-neutral emotions (e.g., \emo{sad} to \emo{anger}, \emo{happy} to \emo{excited}) are more frequent in IEMOCAP than in \DailyDialog{}; see \citep{ghosal2020utterancelevel}. Consequently, both cause detection and causal reasoning in IEMOCAP are more interesting as well as difficult. Lastly, in \cref{tab:stat}, we can see that in our annotated IEMOCAP split, almost 40.5\% of utterances have their emotion cause in utterances at least $3$ timestamps distant in the contextual history. 
% On the contrary, 
In contrast,
this percentage is just $13\%$ in our annotated \DailyDialog{} dataset. 

\subsection{Annotation Process}\label{sec:annot}
% \paragraph{Annotation Task Formulation}

\paragraph{Annotators}
The annotators were undergraduate and graduate computer science students. They had adequate knowledge about the problem of emotion cause recognition; in particular, we organized a special workshop to instruct them on the topic. Their annotations were first verified on a trial dataset, and feedback was provided to them to correct their mistakes. Once they achieved satisfactory performance on the trial dataset, they were qualified for the main dataset annotation. While the annotators were not native English speakers, they communicate in English in their daily life, and their medium of instruction in their study was English.

\paragraph{Annotation guidelines}
Given an utterance $U_t$ labeled with an emotion $E_t$, the annotators were asked to extract the set of causal spans $CS(U_t)$ 
% from the conversational history $H(U_t)$ (including utterance $U_t$) 
that sufficiently represent the causes of the emotion~$E_t$. If the cause of $E_t$ was latent, i.e., there was no explicit causal span in the dialog,
% $H(U_t)$,
% {@SOUJANYA: is the context and history the same? If so, we should either use only one term, or note in \S3.2 that the history is also called context.
% Soujanya: Not same. Context can be both past and future. Historical context is only the past.}
the annotators wrote down the assumed causes that they inferred from the text. Each utterance was annotated by two human experts---graduate students with reasonable knowledge of the task.

\begin{figure*}[t]
     \centering
     \begin{subfigure}[b]{60ex}
         \centering
         \includegraphics[width=\textwidth]{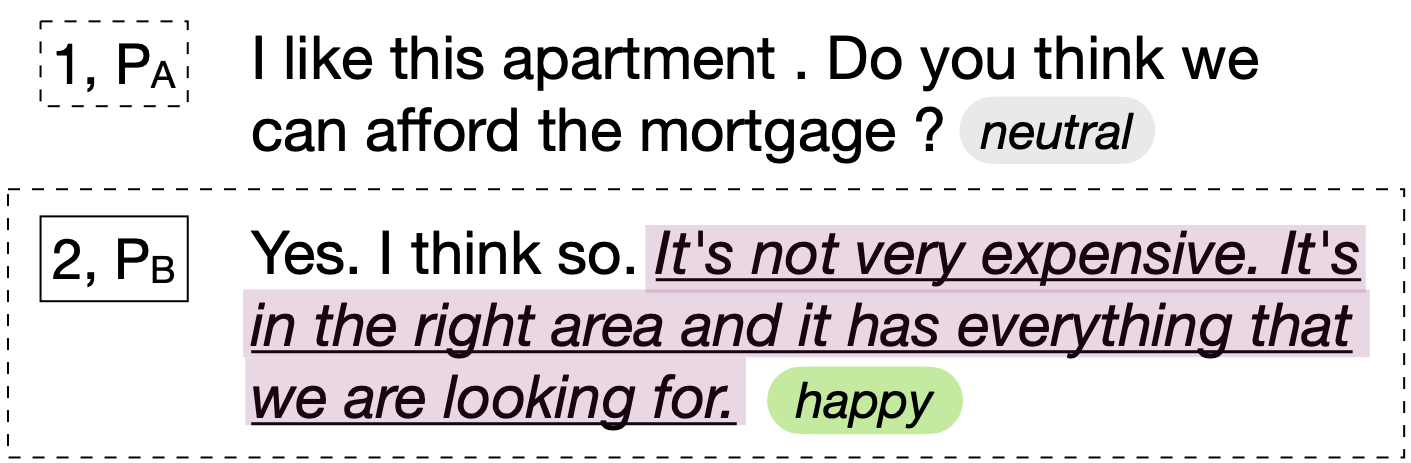}
         \caption{}
         \label{fig:emotion_in_target}
     \end{subfigure}
     % \hfill
     \\[3ex]
     \begin{subfigure}[b]{60ex}
         \centering
         \includegraphics[width=\textwidth]{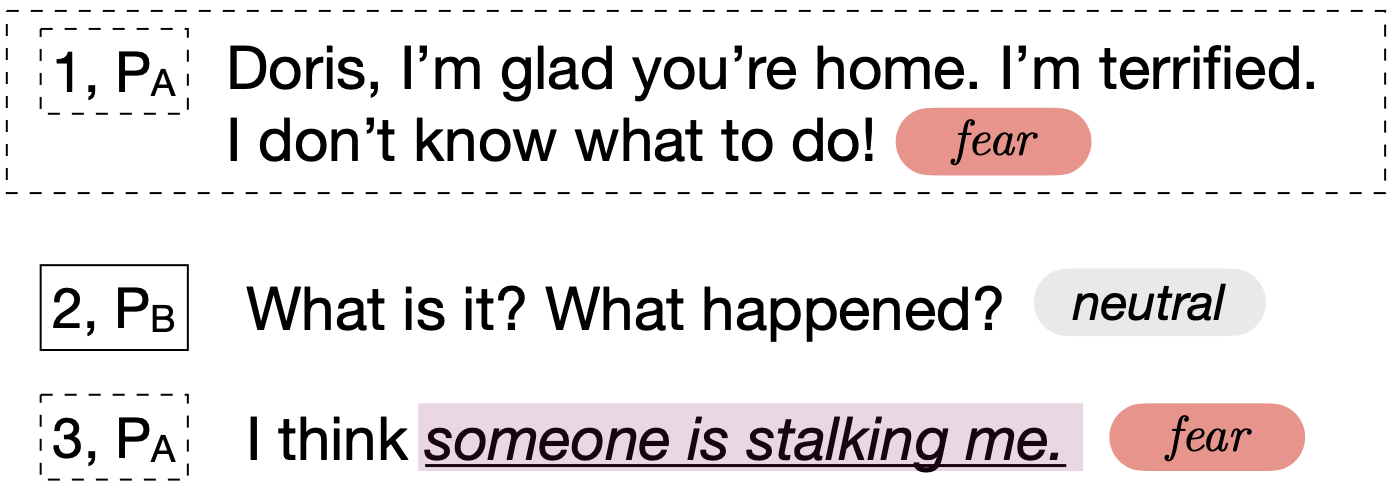}
         \caption{}
         \label{fig:latent_cause}
     \end{subfigure}
     % \hfill
     \\[3ex]
     \begin{subfigure}[b]{60ex}
         \centering
         \includegraphics[width=\textwidth]{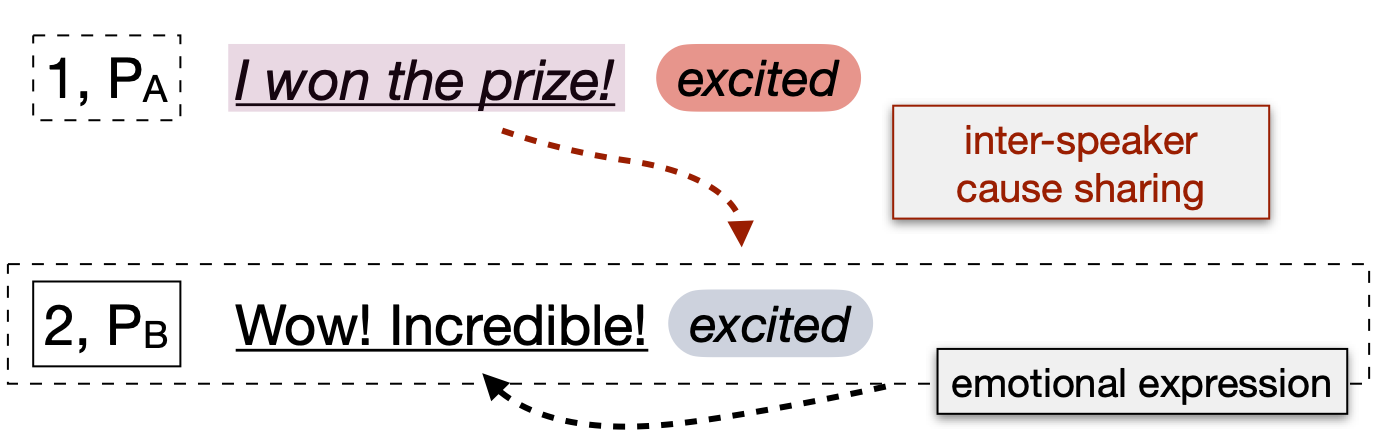}
         \caption{}
         \label{fig:causevsexpression}
     \end{subfigure}
     \caption{{($a$) No context. {($b$)} Unmentioned latent cause. {($c$}) Distinguishing emotion cause from emotional expressions}.}
\end{figure*}

In fact, the annotators were asked to look for the casual spans of $U_t$ in the whole dialog and not only in the past history $H(U_t)$. We show 
a
% one such 
case in \cref{fig:latent_cause} where the causal span of the emotion \emo{fear} in utterance~1 is recognized in utterance~3: ``someone is stalking me''.
% also flagged the utterances with explicit emotion causal spans that occur in the conversational future with respect to $U_t$. 
However, 
% they flagged 
there were
only seven instances of the utterances with explicit emotion causal spans 
% that occur 
in the conversational future with respect to $U_t$ in the whole dataset.
% which is too few for supervised learning. 
% As such, 
So
we discarded those spans and decided to consider only causal spans in $H(U_t)$; hence the definition in~\cref{sec:terminology}.

\paragraph{Emotional expression}
An utterance can contain \textit{$(i)$}~a description of the triggers or stimuli of the expressed emotion, and~/~or \textit{$(ii)$}~a reactionary emotional expression. 
In our setup, by following the discrimination among emotion evidence and cause as explained in~\cref{sec:terminology}, we instructed the annotators to look beyond just emotional expressions and 
% strive for  
% \?{@SOUJANYA: in what sense? If there is at least one trigger in the whole history, then do not label reactionary? Or, the same but by each individual utterance?}
identify the 
% actual 
emotion cause. We can illustrate this with
% one such case in 
\cref{fig:causevsexpression}, where $P_A$ explains the cause for \emo{happiness}; the same cause evokes the emotion \emo{excited} in $P_B$. Meanwhile, the utterance $2$ by $P_B$ is merely an emotional expression (evidence).
% \begin{figure}[t!] 
%     \centering 
%     \includegraphics[width=0.7\columnwidth]{causevsexpression.pdf} 
%     \caption{{Distinguishing emotion cause from emotional expressions.}}
%     \label{fig:causevsexpression}
% \end{figure}

\ToDeleteIfNeedSpace % this is out of topic
Emotion cause can also corroborate in generating an emotional expression, e.g., in \cref{fig:causevsexpression}, the event ``\textit{winning the prize}'' causes \emo{excited} emotion in $P_B$ which directs $P_B$ to utter the expression ``\textit{Wow! Incredible}''. This type of generative reasoning will be very important in our future work.

\paragraph{Why span detection?}
First, emotion-cause extraction has historically been defined as an information extraction task of identifying spans within the emotion-bearing sentences~\cite{DBLP:conf/acl/XiaD19,DBLP:conf/cicling/GhaziIS15}. The core assumption is that such spans are good descriptors of the underlying causes of the generated emotions~\cite{talmy2000toward}. We extend this popular formalism into a multi-span framework.
Second, while recognizing emotion cause is driven by multiple controlling variables such as goal, intent, personality, we adopt this setup as these spans can often represent or allude to these controlling variables. A more elaborate setup would require explaining how the spans can be combined to form the trigger and consequently evoke the emotion (see \cref{fig:csk_exx}); we leave such emotion causal reasoning in conversations to future work.
%\begin{figure}[ht]
%    \centering
%    \includegraphics[width=50ex]{csk.png}
%    \caption{{An example of emotional reasoning where the \emph{happiness} in utterance~3 is caused by the triggers in utterances~1 and~2.}}
%    \label{fig:csk_ex}
%\end{figure}

\subsubsection{Annotation Aggregation}
% We aggregate the annotations in two steps:
% \hl{Follow this \protect\url{https://www.aclweb.org/anthology/D16-1170.pdf}}
Following \citet{gui2016event}, we aggregate the annotations in two stages: 
at utterance and span level.
% utterance-level and span-level aggregation.

\paragraph{Stage 1: Utterance-level aggregation}
Here, we decide whether an utterance is causal by majority voting: a third expert annotator is brought in as the tie breaker.

\begin{table}[t!]
\centering
%\small
\resizebox{1\linewidth}{!}
{\begin{tabular}{@{}l@{}c@{~~~}c@{~~~}cc@{}}
\toprule
\textbf{Dataset} & \textbf{Language} & \textbf{Source} & \textbf{Size} & \textbf{Format}\\
\midrule
\citet{DBLP:conf/ijcnlp/NeviarouskayaA13} & English    & ABBYY Lingvo dictionary  & 532& sentences\\
% &  & dictionary & 532 & Sentences \\
\citet{DBLP:conf/nlpcc/GuiYXLLZ14} & Chinese   & Chinese Weibo & 1333& sentences\\
\citet{DBLP:conf/cicling/GhaziIS15} & English    & FrameNet  & 1519  & sentences\\
\citet{gui2016event} & Chinese  & SINA city news & 2167  & clauses\\
\citet{gao2017overview} & Chinese / Eng. & SINA city news / English novel & 4054 / 4858 & clauses \\
% & English & /English Novel &  2403 & Clauses \\
\midrule
\multirow2*{\RECCONDA{} (our)} & \multirow2*{English} & \multirow2*{\DailyDialog{} / IEMOCAP} & 5861 / 494   & utterances \\
&& & 1106 / 16& dialogues\\
% &  & \DailyDialog{} & 9915 & Dialogues \\
\bottomrule
\end{tabular}}
	\caption{{Datasets for emotion cause extraction and related tasks. Datasets in \cite{DBLP:conf/acl/XiaD19,chen-etal-2020-conditional} are derived from \cite{gui2016event}.}}
	% \hl{Size of \RECCONDA{} is in UCS.}}}
	\label{tab:related_datasets}
\end{table}

\paragraph{Stage 2: Span-level aggregation}
Within each causal utterance selected at 
% the previous step, 
stage 1,
we took the union of the candidate spans from different annotators as the final causal span only when the size of their intersection is at least 50\% of the size of the shortest candidate span. 
% In other words, for overlapping annotated spans, we take the larger boundary as the final causal span.
Otherwise,
%If candidate spans do not share a sub-span or their intersection is 
%% shorter 
%less
%than 50\% of the shortest candidate span, 
a third annotator was brought in to determine the final span from the existing spans. This third annotator was also instructed to prefer the shorter spans over the longer ones when they can sufficiently represent the cause without losing any information. The threshold of 50\% of the shortest span was chosen empirically by examining a small subset of the dialogues.
The third annotator could not break the tie for~34 causal utterances, which we discarded from the dataset.

% \paragraph{Ambiguity in the Annotation Process}
% Emotion Recognition in Conversations (ERC) is a difficult problem without the presence of multimodal sensory inputs i.e., facial expressions and voice of the interlocutors. While annotating the utterances in \DailyDialog{}, our annotators faced issues  

% \hl{Share experience or express opinion, no context}
% \hl{ERC is hard without multimodal inputs. Our annotators reported confounding emotion labels---some utterance may be labeled as both neutral and non-neutral. give examples.}
% % \hl{https://docs.google.com/spreadsheets/d/17dnAdrPwvdaEa4DaVzcVWHgT-ZycIdS_/edit#gid=1849985539}

\begin{figure*}[t]
     \centering
     \begin{subfigure}[b]{60ex}
         \centering
         \includegraphics[width=\textwidth]{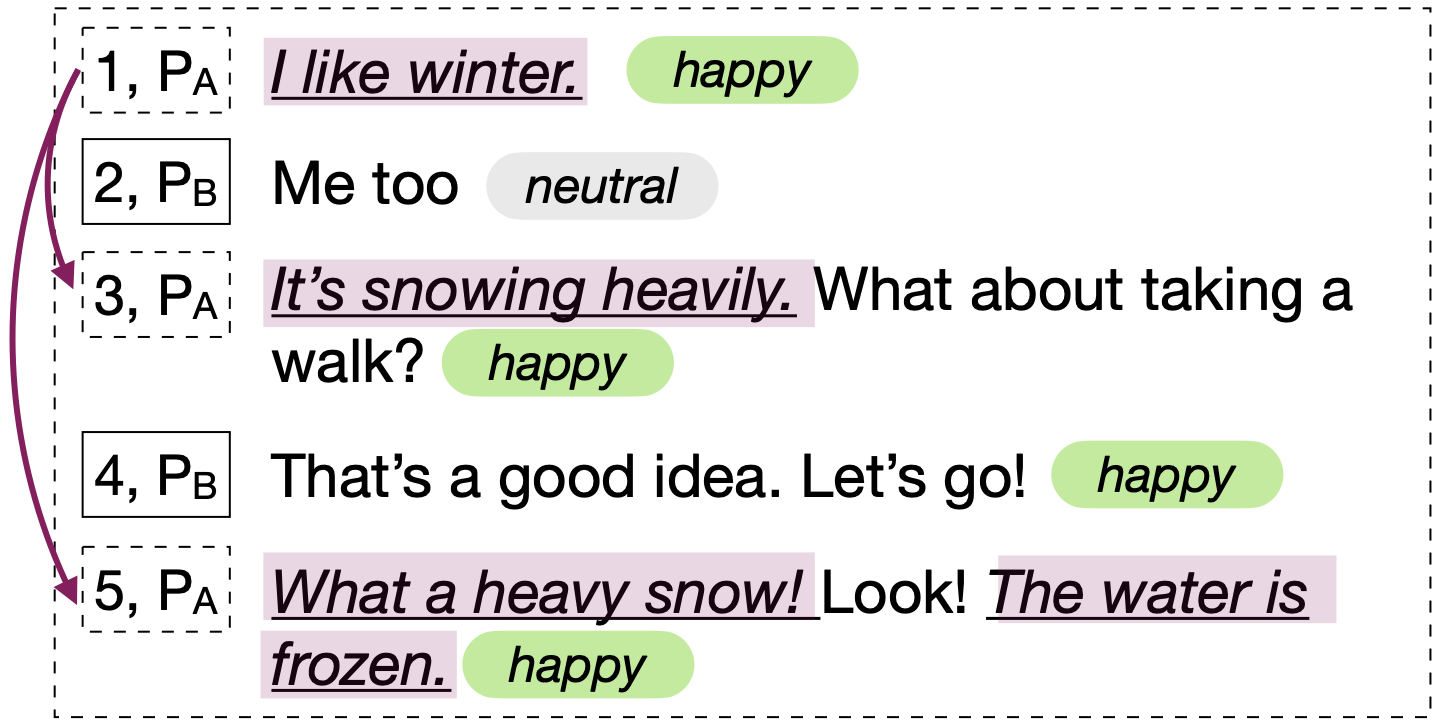}
         \caption{Mood Setting}
         \label{fig:self-contagion}
     \end{subfigure}
     % \hfill
     \\[3ex]
     \begin{subfigure}[b]{60ex}
         \centering
         \includegraphics[width=\textwidth]{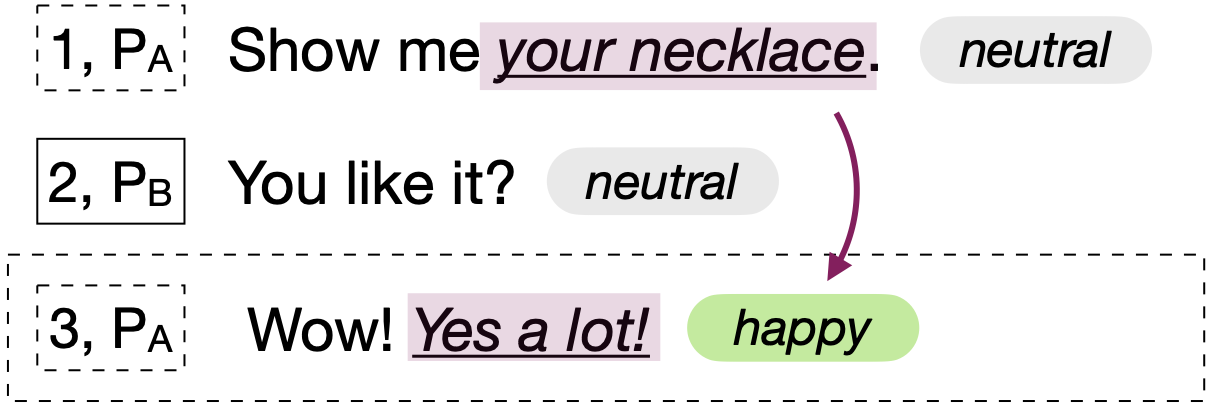}
         \caption{Generic Cause}
         \label{fig:generic_cause}
     \end{subfigure}
     % \hfill
     \\[3ex]
     \begin{subfigure}[b]{60ex}
         \centering
         \includegraphics[width=\textwidth]{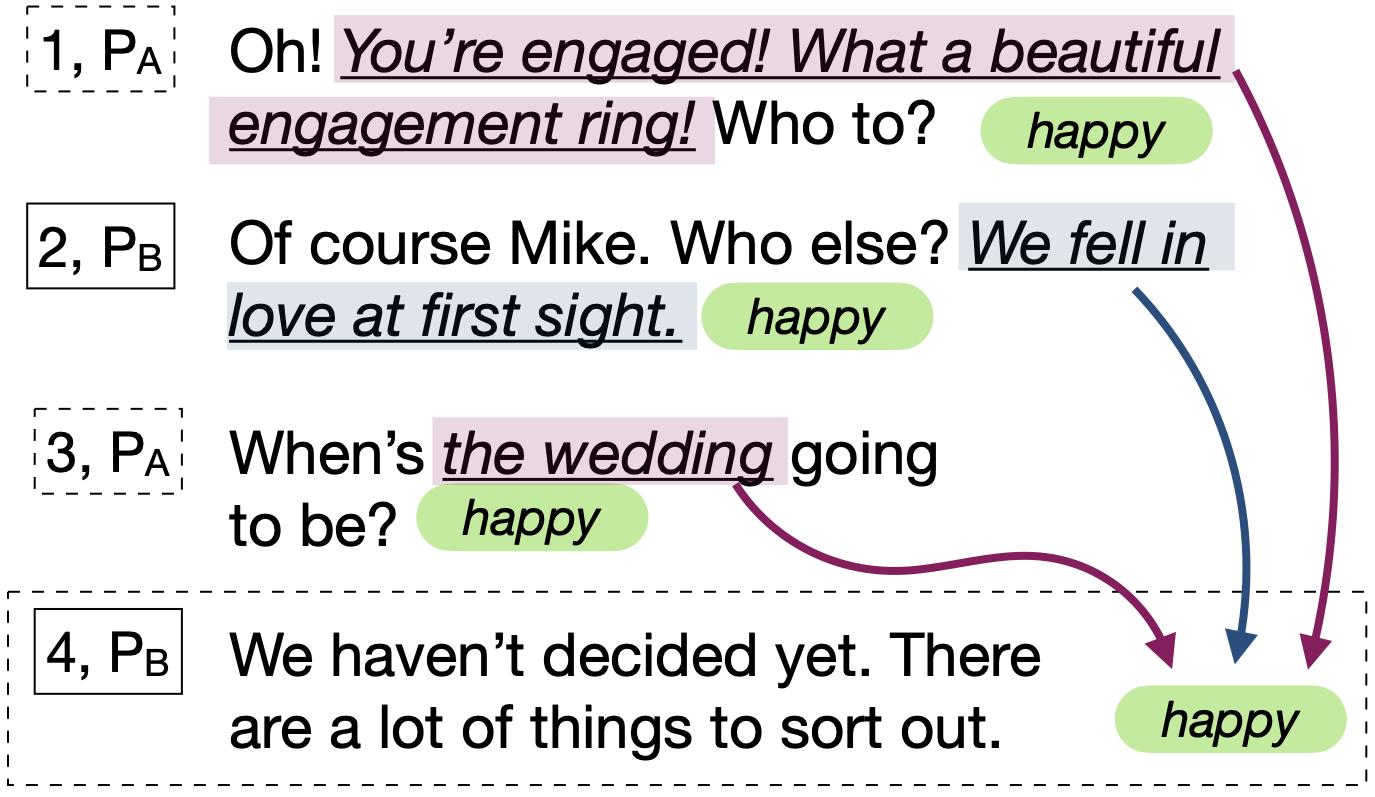}
         \caption{Hybrid}
         \label{fig:hybrid}
     \end{subfigure}
     \caption{{($a$), ($b$) \textit{Self-contagion:} % (\ref{fig:self-contagion},\ref{fig:generic_cause}):} 
     The cause of the emotion is primarily due to a stable mood of the speaker that was induced in the previous dialogue turns; ($c$) \textit{Hybrid:} % (\ref{fig:hybrid}):} 
     The hybrid type 
     % where the role of 
     with both
     inter-personal emotional influence and self-contagion.
     % can be observed.
     }}
\end{figure*}

\begin{table}[t!]
%\small
% \tiny
 	\centering
 	%\resizebox{1\linewidth}{!}
 	{\begin{tabular}{lr@{~~~}r@{~~~~}r}
		\toprule
		\textbf{Number of items} & \multicolumn1c{\textbf{DD}}  & \multicolumn1c{\textbf{IE}} & \textbf{Total}\\
		\midrule
		Dialogues & $1106$ & $16$ & 1122\\
		Utterances & $11104$ & $665$ & 11769\\
		Utterances annotated with emotion cause  & $5861$ & $494$ & 6355\\
		%\# Utterances Annotated with Emotion Cause  & $5861$ & \\
		Utterances that cater to background 
		% evidence 
		cause & $395$ & $70$ & 465\\
		Utterances where cause solely lies in the same utterance & $1521$ & $80$ & 1601\\
		Utterances where cause solely lies in the contextual utterances & $970$ & $171$ & 1141\\
		Utterances where cause lies both in same and context utterances 
		% along with contextual utterances 
		& $3370$ & $243$ & 3613\\
%		\midrule
%		Utterances with emotion \emo{Anger}       & 451 & 89 & 540\\
%        Utterances with emotion \emo{Fear}        & 74 & -- &74\\
%        Utterances with emotion \emo{Disgust}     & 140 & -- &140\\
%        Utterances with emotion \emo{Frustration} &  -- & 109 &109\\
%        Utterances with emotion \emo{Happy}       & 4361 & 58 &4419\\
%        Utterances with emotion \emo{Sad}         & 351 & 70 &4419\\
%        Utterances with emotion \emo{Surprise}    & 484 & -- &484\\
%        Utterances with emotion \emo{Excited}     & -- & 197 &197\\
%        Utterances with emotion \emo{Neutral}     & 5243 & 142 &5385\\  
		\midrule
		UCS pairs & $9915$ & $1154$ &11069\\
		Utterances having single cause & $55\%$ & $41\%$ & 54\%\\
		Utterances having two causes & $31\%$ & $24\%$ & 31\%\\
		Utterances having three causes & $9\%$ & $17\%$ & 9\%\\
		Utterances having more than three causes & $5\%$ & $18\%$ & 6\%\\
		Causes per utterance (average) & $1.69$ & $2.34$ &1.73\\
%		\midrule
%		No Context & $43\%$ & $35\%$ &43\%\\
%		Inter-Personal & $32\%$ & $19\%$ &31\%\\
%		Self-Contagion & $9\%$ & $20\%$ &10\%\\
%		Hybrid & $11\%$ & $17\%$ &11\%\\
%		Latent & $5\%$ & $10\%$ &5\%\\
%		\midrule
%		Utterances ($U_t$) having cause at $U_{(t-1)}$ & $2851$ & $183$ &3034\\
%		Utterances ($U_t$) having cause at $U_{(t-2)}$ & $1182$ & $124$ & 1306\\
%		Utterances ($U_t$) having cause at $U_{(t-3)}$ & $578$ & $94$ &672\\
%		Utterances ($U_t$) having cause at $>U_{(t-3)}$ & $769$ & $200$ &969\\
		\bottomrule
	\multicolumn4{@{}c@{}}{\begin{tabular}{@{}c@{}}
	    \begin{tabular}[t]{lr@{~~~}r@{~~~~}r}
	    \toprule
		\textbf{Utterances with} & \multicolumn1c{\textbf{DD}}  & \multicolumn1c{\textbf{IE}} & \textbf{Total}\\
		\midrule
		 \emo{Anger}       & 451 & 89 & 540\\
         \emo{Fear}        & 74 & -- &74\\
         \emo{Disgust}     & 140 & -- &140\\
         \emo{Frustration} &  -- & 109 &109\\
         \emo{Happy}       & 4361 & 58 &4419\\
         \emo{Sad}         & 351 & 70 &4419\\
         \emo{Surprise}    & 484 & -- &484\\
         \emo{Excited}     & -- & 197 &197\\
         \emo{Neutral}     & 5243 & 142 &5385\\  
        \bottomrule
	    \end{tabular} ~~~~ \begin{tabular}[t]{lr@{~~~}r@{~~~~}r}
	    \toprule
		\textbf{Utterances $U_t$ with} & \multicolumn1c{\textbf{DD}}  & \multicolumn1c{\textbf{IE}} & \textbf{Total}\\
		\midrule
		No context & $43\%$ & $35\%$ &43\%\\
		Inter-personal & $32\%$ & $19\%$ &31\%\\
		Self-contagion & $9\%$ & $20\%$ &10\%\\
		Hybrid & $11\%$ & $17\%$ &11\%\\
		Latent & $5\%$ & $10\%$ &5\%\\
		\midrule
		Cause at $U_{(t-1)}$ & $2851$ & $183$ &3034\\
		Cause at $U_{(t-2)}$ & $1182$ & $124$ & 1306\\
		Cause at $U_{(t-3)}$ & $578$ & $94$ &672\\
		Cause at $U_{(t-\ge4)}$ & $769$ & $200$ &969\\
		\bottomrule
	    \end{tabular}
	\end{tabular}}
	\end{tabular}}
	\caption{{Statistics of the \RECCONDA{} annotated dataset. DD stands for \RECCONDADD{}, IE for \RECCONDAIE{}.}}
	\label{tab:stat}
\end{table}

\subsection{Dataset Statistics}
In \cref{tab:related_datasets}, we compare our dataset with the existing datasets in terms of size, data sources, and language. The remaining statistics of \RECCONDA{} are consolidated in \cref{tab:stat}.

We measured 
% two types of inter-annotator agreement scores: \textit{$(i)$}~at the utterance level and  \textit{$(ii)$}~at the span level.
inter-annotator agreement (IAA) at the level of $(i)$~utterance and $(ii)$~span.
At the utterance level, we measured IAA following \citet{gui2016event}, which gave a kappa of 0.7928. 
% By adopting the macro F1 score scheme as pointed by \citet{brandsen-etal-2020-creating}, we get a 0.8201 macro F1 score. In this scheme, we discard the utterances in the conversational history containing no causal span for the emotion of the target utterance as they are very frequent and consequently may lead to a skewed F1 score. At span level, the F1 score, as explained in \citet{rajpurkar2016squad}, is calculated for all possible pairs of annotators followed by taking their average. Overall, we obtain an F1 score of 0.8035 at the span level.
However, as pointed out by \citet{brandsen-etal-2020-creating}, macro F1 score is 
% a 
more appropriate 
% approach 
for span extraction-type tasks. Hence, at the utterance level, we also compute the pairwise macro F1 score between all possible pairs of annotators and then average them, which 
% This gives us 
gives
a 0.8839 macro F1 score. \citet{brandsen-etal-2020-creating} also suggest the removing negative examples---in our case, the utterances in the conversational history containing no causal span for the emotion of the target utterance---for macro F1 calculation, since such examples are usually very frequent, which may lead to a skewed F1 score. As expected, 
% adopting 
this yields a lower F1 score of 0.8201. At span level, the F1 score, as explained in \citet{rajpurkar2016squad}, is calculated for all possible pairs of annotators followed by taking their average. Overall, we obtain an F1 score of 0.8035 at span level.

\section{Types of Emotion Causes}
\label{sec:types}
In our dataset, \RECCONDA{}, we observe five predominant types of emotion causes that are based on the source of the stimuli (events / situations / acts) in the conversational context, responsible for the target emotion. 
% These cases require different kinds and amounts of context to identify the respective spans which indicate the stimuli (events/situations/acts) responsible for the target emotion.
The annotators were asked to flag the utterances with latent emotion cause or emotion cause of type 
% 2b, 
shown in \cref{fig:latent_cause} (unmentioned latent cause),
as explained below.
% For the rest of the types, utterance ids and speaker information are used to determine the emotion-cause types. 
The distribution of these cause types is given in \cref{tab:stat}.

% \?{COMMENT:} specify type 2b (Unmentioned Latent Cause). It can be confused with the items handled in Type 2.
% \?? \?{@GELBUKH she is right: it is unmentioned latent cause; check the first sentence in figure 2.b; "I'm terrified. I dont know what to do"}

\paragraph{Type 1: No Context} The cause is present within the target utterance itself. The speaker feeling the emotion explicitly mentions its cause in the target utterance (see~\cref{fig:emotion_in_target}).
 
% \begin{figure}[t!]
%     \centering
%     \includegraphics[width=0.7\columnwidth]{target_utt_cause.pdf}
%     \caption{{In utterance $2$, the cause is mentioned by the speaker within the utterance, thus not requiring the need for context.}}
%     \label{fig:emotion_in_target}
% \end{figure}

\paragraph{Type 2: Inter-Personal Emotional Influence}
The emotion cause is present in the other speaker's utterances (see \cref{fig:examples}).
% We observe nuanced differences between possible emotion causes between dyadic speaker turns.
We observe two possible sub-types of such influences:
\begin{enumerate}[itemsep=0ex, leftmargin=*, label=2\alph*)]
    \item \textbf{Trigger Events / Situations.} The emotion cause lies within an event or concept mentioned by the other speaker.
    \item \textbf{Emotional Dependency.} The emotion of the target speaker is induced from the emotion of the other speaker over some event / situation.
\end{enumerate}

\paragraph{Type 3: Self-Contagion}
In many cases, 
% we observe that 
the cause of the emotion is primarily due to a stable mood of the speaker that was induced in 
% some 
previous dialogue turns. 
% For example, 
E.g.,
in a dialogue involving cordial greetings, there is a tendency for a \emo{happy} mood to persist across several turns for a speaker. \cref{fig:self-contagion} presents an example where such self-influences can be observed. Utterance $1$ establishes that $P_A$ likes winter. This concept triggers a \emo{happy} mood for the future utterances, as observed in utterances~3 and~5. In \cref{fig:generic_cause}, similarly, the trigger of emotion \emo{excited} in utterance~3 is mentioned by the same speaker in his or her previous utterance.

\paragraph{Type 4: Hybrid} Emotion causes of type~2 and~3 can jointly cause the emotion of an utterance, as illustrated by \cref{fig:hybrid}.

% \begin{figure}[ht!]
% 	\centering
% 	\includegraphics[width=0.7\linewidth]{latent.pdf}
%  	\caption{{Unmentioned Latent Cause.}}
%  	\label{fig:latent_cause}
% \end{figure}

\paragraph{Type 5: Unmentioned Latent Cause} There are instances in the dataset where no explicit span in the target utterance or the conversational history can be identified as the emotion cause. \cref{fig:latent_cause} shows such a case. Here, in the first utterance, $P_A$ speaks of being terrified and fearful without indicating the cause. We annotate such cases as latent causes. Sometimes the cause is revealed in future utterances, e.g., ``\textit{someone is stalking me}'' as the reason of being fearful. However, as online settings would not have access to the future turns, we refrain from treating future spans as 
% causal evidence.
causes.

\section{Experiments}
\label{sec:experiments}

We formulate two distinct subtasks of \RECCON{}: \textit{$(i)$}~causal span extraction and \textit{$(ii)$}~causal emotion entailment. However, 
% it should be noted that 
note that
the main purposes of this work are to present a dataset and setup the strong baselines.

\subsection{Compiling Dataset Splits}
\label{sec:dataprep}

\RECCONDADD{} is the subset of our dataset that contains dialogues from \DailyDialog{}. For this subset, we created the training, validation, and testing examples based on the original splits in~\cite{li2017DailyDialog}. However, this resulted in the validation and testing sets to be quite small, so we moved some dialogues to them from the original training set. 

The subset \RECCONDAIE{} consists of dialogues from the IEMOCAP dataset. This subset is quite small as it contains only sixteen unique dialogues (situations). So, we consider the entire \RECCONDAIE{}  as another testing set, emulating an out-of-distribution generalization test. We report results on this dataset based on models trained on \RECCONDADD{}. In our experiments, we ignore the utterances with only latent emotion causes. 

\subsubsection{Generating Negative Examples}
\label{sec:neg}
The annotated dataset, \RECCONDA{} (consisting of subsets \RECCONDADD{} and \RECCONDAIE{}) only contains positive examples, where an emotion-containing target utterance is annotated with a causal span extracted from its conversational historical context. However, to train a model for the \RECCON{} task, we need negative examples, i.e., the instances which are not cause of the utterance. In the sequel, we use the terminology introduced in \cref{sec:terminology}; the reader should 
% We urge readers to 
refer to that section for clearer understanding.

% To this end, 
% We adopt the following strategy to create the negative examples:
We use three different strategies to create the negative examples.
In this section, we will discuss in detail Fold 1. Then, in \cref{sec:analysisx}, to further analyze the performance of our models, besides Fold 1, we will adopt two more strategies, Fold 2 and Fold 3, to create the negative examples:
% \footnote{For Fold 2 and Fold 3, see~\cref{sec:analysisx}.}
\begin{description}
% [labelindent=0pt,labelwidth=\widthof{\textbf{Fold 3:}},leftmargin=!]
%[leftmargin=*]
\item[\textbf{Fold 1:}] Consider a dialogue~$D$ and a target utterance~$U_t$ in~$D$.
% , and the corresponding causal utterance set $C(U_t)$. 
We construct the complete set of negative examples as 
% $\{(U_t, U^t_n) \mid \forall U^t_n \in H_{U_t} \setminus C_{U_t} \}$. 
$\{(U_t, U_i) \mid U_i \in H(U_t) \setminus C(U_t) \}$,
where~$H(U_t)$ is the conversational history and~$C(U_t)$ is the set of causal utterances for~$U_t$.
\item[\textbf{Fold 2:}] In this scheme, we randomly sample the non-causal utterance~$U_i$ along with the corresponding historical conversational context~$H(U_i)$ from another dialogue in the dataset to create a negative example.
\item[\textbf{Fold 3:}] This is similar to Fold~2 with a constraint. In this case, a non-causal utterance~$U_i$ along with its historical conversational context~$H(U_i)$ from the other dialogue is only sampled when its emotion matches the emotion of the target utterance~$U_t$ to construct a negative example.
\end{description}
%Note that Fold 1 is a more challenging and practical choice compared to the rest of the two folds as in real scenarios, we need to identify causes of emotions within a single dialogue by reasoning over the utterances in it. The statistic of the final dataset is shown in \cref{tab:finalstat}. The valid UCS pairs correspond to the positive examples. The invalid UCS pairs refer to the negative examples. 
Note that unlike Fold 1, a negative example in Fold 2 and 3 comprising a non-causal utterance $U_i$ and a target utterance $U_t$ belong to different dialogues. For the cases where the causal spans do not lie in the target utterance, we remove the target utterance from its historical context when creating a positive example in Fold 2 and 3. As a result, it helps to prevent the models from learning any trivial patterns. The statistics for the three folds are shown in \cref{tab:finalstatx}.

\begin{table}[ht!]
%\small
\centering
% \scalebox{0.7}%
{
%\resizebox{\linewidth}{!}{
\begin{tabular}{llrrrr}
\toprule
& &  \multicolumn1c{Data} & \multicolumn1c{Train} & \multicolumn1c{Val} & \multicolumn1c{Test} \\
\midrule
\multirow{4}{*}{\rotatebox{90}{\textbf{\small{Fold 1}}}} & \multirow{2}{*}{\rotatebox{90}{\textbf{{DD}}}} & Positive UCS pairs & 7269 & 347 & 1894 \\
&& Negative UCS pairs & 20646 & 838 & 5330 \\
\cmidrule{2-6}
&\multirow{2}{*}{\rotatebox{90}{\textbf{{IE}}}} & Positive UCS pairs & \multicolumn1c{--} & \multicolumn1c{--} & 1080 \\
&& Negative UCS pairs & \multicolumn1c{--} & \multicolumn1c{--} & 11305 \\

\midrule
\multirow{4}{*}{\rotatebox{90}{\textbf{\small{Fold 2}}}} & \multirow{2}{*}{\rotatebox{90}{\textbf{{DD}}}} & Positive UCS pairs & 7269 & 347 & 1894 \\
&& Negative UCS pairs & 18428 & 800 & 4396 \\
\cmidrule{2-6}
&\multirow{2}{*}{\rotatebox{90}{\textbf{{IE}}}} & Positive UCS pairs & \multicolumn1c{--} & \multicolumn1c{--} & 1080 \\
&& Negative UCS pairs & \multicolumn1c{--} & \multicolumn1c{--} & 7410 \\

\midrule
\multirow{4}{*}{\rotatebox{90}{\textbf{\small{Fold 3}}}} & \multirow{2}{*}{\rotatebox{90}{\textbf{{DD}}}} & Positive UCS pairs & 7269 & 347 & 1894 \\
&& Negative UCS pairs & 18428 & 800 & 4396 \\
\cmidrule{2-6}
&\multirow{2}{*}{\rotatebox{90}{\textbf{{IE}}}} & Positive UCS pairs & \multicolumn1c{--} & \multicolumn1c{--} & 1080 \\
&& Negative UCS pairs & \multicolumn1c{--} & \multicolumn1c{--} & 7410 \\
\bottomrule
\end{tabular}
%}
}
\caption{{The statistics of \RECCONDA{} comprising both positive (valid) and negative (invalid) UCS pairs. DD stands for \RECCONDADD{}, IE for \RECCONDAIE{}. Utterances with only latent emotion causes are ignored in our experiments.}}
\label{tab:finalstatx}
\end{table}

\subsection{Subtask 1: Causal Span Extraction}
\label{sec:cse}
\textit{Causal Span Extraction} is the task of identifying the causal span (emotion cause) for a target non-neutral utterance. In our experimental setup, we formulate \textit{Causal Span Extraction} as a Machine Reading Comprehension (MRC) task similar to the task in Stanford Question Answering Dataset~\citep{rajpurkar2016squad}. Similar MRC techniques have been used in literature for various NLP tasks such as named entity recognition~\citep{li-etal-2020-unified} and zero shot relation extraction~\citep{levy-etal-2017-zero}. In this work, we propose two different span extraction settings: \textit{$(i)$}~with conversational context and \textit{$(ii)$}~without conversational context.

\subsubsection{Subtask Description}

\paragraph{With Conversational Context (w/ CC)}
We 
% speculate 
believe
that the presence of conversational context would be key to the span extraction algorithms. To evaluate this hypothesis, we design this subtask, where the conversational history is available to the model. In this setup, for a target utterance $U_t$, the causal utterance $U_i \in C(U_t)$, and a causal span $S \in CS(U_t)$ from $U_i$, we construct the context, question, and answer as follows:\footnote{By ``causal span from evidence in the context'' we mean a causal span from the conversation history~$H(U_t)$.}
\begin{description}
%[labelindent=0pt,labelwidth=\widthof{\textbf{Question:}},leftmargin=!]
\item[\textbf{Context:}] The context of a target utterance $U_t$ is the conversational history, i.e., a concatenation of all utterances from $H(U_t)$.
% ) of a target utterance $U_t$ including the target utterance. 
Similarly, for a negative example $(U_t, U_i)$, where $U_i \notin C(U_t)$, conversational history of $U_t$ is used as context.\\
\item[\textbf{Question:}] The question is framed as follows: ``\textit{The target utterance is ${<}U_t{>}$. The evidence utterance is ${<}U_i{>}$. What is the causal span from evidence in the context that is relevant to the target utterance's emotion ${<}E_t{>}$?}".
\item[\textbf{Answer:}] The causal span $S \in CS(U_t)$ appearing in $U_i$ if $U_i \in C(U_t)$. For negative examples, $S$ is assigned an empty string. 
\end{description}

% We employ the same strategy described earlier to handle target utterances having multiple evidence utterances. 
If a target utterance has multiple causal utterances and causal spans, then we create separate (Context, Question, Answer) instances for them. Unanswerable questions are also created from invalid (cause, utterance) pairs following the same approaches explained in \cref{sec:dataprep}. 

\paragraph{Without Conversational Context (w/o CC)}
In this formulation, we intend to identify whether the \textit{Causal Span Extraction} task is feasible when we only have information about the target utterance and the causal utterance. Given a target utterance $U_t$ with emotion label $E_t$, its causal utterance 
% $U_i$ where 
$U_i \in C(U_t)$, and the causal span $S \in CS(U_t)$, the question is framed as framed as follows: ``\textit{The target utterance is ${<}U_t{>}$. What is the causal span from context that is relevant to the target utterance's emotion ${<}E_t{>}$?}''. The task is to extract answer $S \in CS(U_t)$ from context $U_i$. For negative examples, $S$ is assigned an empty string.

\subsubsection{Models}
We use 
% the following 
two pretrained Transformer-based models to benchmark the \textit{Causal Span Extraction} task.
% Two different family of Transformer models are used which are as follows:

\paragraph{{RoBERTa Base}} We use the \code{roberta-base}
% checkpoint of RoBERTa 
model~\cite{liu2019roberta} and add a linear layer on top of the hidden-states output to compute span start and end logits. Scores of candidate spans are computed following~\citet{devlin2018bert}, and the span with maximum score is selected as the answer.
% The scores of each word (in the context) being the start of the span and end of the span are then used to compute the score of a candidate span. All possible candidate spans are scored are the maximum scoring one is selected as the answer.
% \paragraph{RoBERTa Fine-tuned on SQuAD}: The RoBERTa model fine-tuned on the SQuAD 2.0 dataset is used as the second baseline model. We use the \code{roberta-base-squad2} checkpoint. We use the SQuAD fine-tuned version to evaluate how much difference does it make when we use a model that has already been trained on a large question answering dataset.
% This model achieves an exact match score of 79.97\% on the SQuAD 2.0 dataset.

\paragraph{{SpanBERT Fine-tuned on SQuAD}} We use SpanBERT~\citep{joshi2020spanbert} as the second baseline model. 
% in our experimental study. 
SpanBERT follows a different pre-training objective compared to RoBERTa (e.g. predicting masked contiguous spans instead of tokens) and performs better on question answering tasks. 
% The large version of the SpanBERT model consistently outperforms BERT based models on span selection tasks including question answering. However, 
In this work we are using the SpanBERT base model fine-tuned on SQuAD 2.0 dataset.
%~\citep{rajpurkar2018know}. 

%The checkpoint name is \code{spanbert-finetuned-squadv2}.

\subsubsection{Evaluation Metrics}
\label{sec:metric}
% Let us define:\\
% $\tau$ = $\#of\_exactly\_matched\_spans$,\\
% %$\alpha$ = $ \# of\_partially\_matched\_spans$,\\
% $T$ = $\#of\_total\_ground\_truth\_spans$.
%We report the following metrics over the positive (Context, Question, Answer) samples. negative examples are excluded because those can be created in any arbitrary number which would skew the results. 
%

We use the following evaluation metrics.
\textbf{EM$_{Pos}$ (Exact Match):} EM represents, with respect to the gold standard data, how many causal spans are exactly extracted by the model.
%\\
% \textbf{PM (Partial Match):} PM represents how many extracted spans partially match the spans in the gold standard data---$ PM = \dfrac{\alpha - \tau}{T}$. For each ground truth span, we increase $\alpha$ by $1$ if at least one word from the ground truth span is found in the predicted span.\\ 
% \textbf{NM (No Match):} NM refers to the case of no matching is found. It can be represented as---$NM = 1 - EM - PM$.\\
% \textbf{{Pos$_{LCS_{F1}}$:}} We compute a F1 score over the longest common sub-sequence of words between the predicted and ground span. This metric is computed for every positive (Context, Question, Answer) instance and then averaged in the dataset level. \\
% The $LCS_{F1}$ measure is defined as follows: 
% \begin{flalign}
%  LCS_{P} &= \dfrac{L(predicted\_span, exact\_span)}{F(predicted\_span)}\\
%  LCS_{R} &= \dfrac{L(predicted\_span, exact\_span)}{F(exact\_span)}\\
%  LCS_{F1} &= \dfrac{2 \times LCS_P \times LCS_R}{LCS_P + LCS_R}
% \end{flalign}
% The function $L$ produces the number of words in the longest common sub-sequence between the predicted and ground span. The function $F$ yields the number of words in a span. This metric is computed for every ground truth span and then averaged in the dataset level.\\
\textbf{F1$_{Pos}$}:~This is the F1 score introduced by~\citet{rajpurkar2016squad} to evaluate predictions of extractive QA models and calculated over positive examples in the data.
%\\ 
%Unlike $Pos_{LCS_{F1}}$, this metric does not consider sequence information.\\
% \textbf{F1:} This is similar to the F1 score explained in~\cite{rajpurkar2016squad}. Here, both ground truth and the predicted spans are represented as bag of tokens and then the F1 score is computed for each span which we then average in the dataset level to obtain the final F1 score. 
% \textbf{Temperature ($\tau$):} We calculate the metric temperature as follows---$\tau = EM + PM \times LCS_{PM}$. The range of $\tau$ is [0,1] where a score of 1 is considered as the best possible outcome. This metric can be used to compare different models.\\
\textbf{F1$_{Neg}$}:~Negative F1 represents the F1 score of detecting negative examples with respect to the gold standard data. Here, for a target utterance $U_t$, the ground truth are empty spans.
%\\
\textbf{F$_1$}:~This metric is similar to F1$_{Pos}$ but calculated for every positive and negative example followed by an average over them.
% \textbf{1 - non-Recall:} This metric indicates the number of empty spans for which non-empty spans are extracted by models.

\begin{table}[t]
  \centering
 % \small
 %\tiny
% \resizebox{\linewidth}{!}
%\scalebox{0.7}{
  {\setlength{\tabcolsep}{1ex}\begin{tabular}{lllccccccccc}
    \toprule
   \multicolumn3c{\multirow3*{\textbf{Model}}} & \multicolumn{4}{c}{\textbf{w/o CC}} && \multicolumn{4}{c}{\textbf{w/ CC}}\\
    \cmidrule{4-7}
    \cmidrule{9-12}
   & & & EM$_{Pos}$ & F1$_{Pos}$ & F1$_{Neg}$ & $F_1$ && EM$_{Pos}$ & F1$_{Pos}$  & F1$_{Neg}$ & $F_1$ \\
    \midrule
   \multirow{4}{*}{\rotatebox{90}{\textbf{{Fold 1}}}} & \multirow{2}{*}{\rotatebox{90}{\textbf{{DD}}}} & RoBERTa  & 26.82 & 45.99 & \textbf{84.55} & \textbf{73.82} && 32.63 & 58.17 & 85.85 & 75.45\\
   
  &  & SpanBERT & \textbf{33.26} & \textbf{57.03} & 80.03 & 69.78 && \textbf{34.64} & \textbf{60.00} & \textbf{86.02} & \textbf{75.71} \\
    \cmidrule{2-12}
    & \multirow{2}{*}{\rotatebox{90}{\textbf{{IE}}}} & RoBERTa  & \09.81 & 18.59 & \textbf{93.45} & \textbf{87.60} && 10.19 & 26.88 & \textbf{91.68} & \textbf{84.52}\\
    
  &  & SpanBERT & \textbf{16.20} & \textbf{30.22} & 87.15 & 77.45 && \textbf{22.41}  & \textbf{37.80} & 90.54 & 82.86 \\
 
      \midrule
      % Fold 1 - UCS_c/1
   \multirow{4}{*}{\rotatebox{90}{\textbf{{$\overline{\text{Fold 1}}$}}}} & \multirow{2}{*}{\rotatebox{90}{\textbf{{DD}}}} & RoBERTa  & 37.76 & 63.87 & -- & -- && 39.02 & 69.13 & -- & --\\
   
  &  & SpanBERT & 41.96 & 72.01 & -- & -- && 42.24 & 71.91 & -- & --\\
    \cmidrule{2-12}
    & \multirow{2}{*}{\rotatebox{90}{\textbf{{IE}}}} & RoBERTa  & 22.49 & 45.01 & -- & -- && 17.27 & 42.15 & -- & --\\
    
  &  & SpanBERT & 26.91 & 52.22 & -- & -- && 31.33 & 60.14 & -- & --\\
    \bottomrule
   \end{tabular}}
%   }
  \caption{{Results for Causal Span Extraction task on the test sets of \RECCONDADD{} and \RECCONDAIE{}. All scores are in percentage and are reported at best validation F1 scores. DD 
  % $\xrightarrow{}$ 
  stands for
  \RECCONDADD{}, IE 
  % $\xrightarrow{}$ 
  for \RECCONDAIE{}, 
  RoBERTa 
  % $\xrightarrow{}$ 
  for
  RoBERTa Base. For definition of Fold 1, see \cref{sec:neg}.}}
  \label{tab:cse}
\end{table}

While all 
% the above 
these
metrics are important for evaluation, we stress that future works should 
particularly consider performances for EM$_{Pos}$, F1$_{Pos}$, and F$_1$.
% For the future models to compare with the baselines introduced in this work, EM, $LCS_{F1}$ and F1 should be reported as the metrics.

\begin{table}[t]
  \centering
 % \tiny
%  \small
% \tabcolsep=3pt
%\scalebox{0.6}{
%  \resizebox{\linewidth}{!}
 {
 \begin{tabular}{@{}lllccccccc@{}}
    \toprule
      \multicolumn{3}{c}{\multirow3*{\textbf{Model}}} & \multicolumn{3}{c}{\textbf{w/o CC}} && \multicolumn{3}{c}{\textbf{w/ CC}}\\
    \cmidrule{4-6}
    \cmidrule{8-10}
    && & Pos. F1 & Neg. F1 & macro F1 && Pos. F1 & Neg. F1 & macro F1\\

    \midrule
    \multirow{10}{*}{\rotatebox{90}{\textbf{{Fold 1}}}} & \multirow{5}{*}{\rotatebox{90}{\textbf{{DD}}}} &  Base & \textbf{56.64} & 85.13 & \textbf{70.88} && 64.28 & 88.74 & 76.51 \\
   & &  Large & 50.48 & \textbf{87.35} & 68.91 && \textbf{66.23} & 87.89 & \textbf{77.06} \\
   & &  ECPE-MLL & -- & -- & -- && 48.48 & 94.68 & 71.59 \\
   & &  ECPE-2D & -- & -- & -- && 55.50 & 94.96 & 75.23 \\
   & &  RankCP & -- & -- & -- && 33.00 & \textbf{97.30} & 65.15 \\
    % &  MNLI & 55.19 & 87.95 & 71.57 & & & \\
    \cmidrule{2-10}
   & \multirow{5}{*}{\rotatebox{90}{\textbf{{IE}}}} &  Base & 25.98 & 90.73 & 58.36 && 28.02 & 95.67 & 61.85\\
  &  &  Large & \textbf{32.34} & \textbf{95.61} & \textbf{63.97} && \textbf{40.83} & \textbf{95.68} & \textbf{68.26} \\
     & &  ECPE-MLL & -- & -- & -- && 20.23 & 93.55 & 57.65 \\
   & &  ECPE-2D & -- & -- & -- && 28.67 & 97.39 & 63.03 \\
    & &  RankCP & -- & -- & -- && 15.12 & 92.24 & 54.75 \\
  
    \midrule
    \multirow{10}{*}{\rotatebox{90}{\textbf{{$\overline{\text{Fold 1}}$}}}} & \multirow{5}{*}{\rotatebox{90}{\textbf{{DD}}}} &  Base & 93.12 & -- & -- && 92.64 & -- & -- \\
   & &  Large & \textbf{98.87} & -- & -- && \textbf{97.78} & -- & -- \\
    % &  MNLI & 55.19 & 87.95 & 71.57 & & & \\
       & &  ECPE-MLL & -- & -- & -- && 84.50 & -- & -- \\
   & &  ECPE-2D & -- & -- & -- && 88.13 & -- & -- \\
   & &  RankCP & -- & -- & -- && 85.67 & -- & -- \\
    \cmidrule{2-10}
   & \multirow{5}{*}{\rotatebox{90}{\textbf{{IE}}}} &  Base & 71.98 & -- & -- && 58.52 & -- & -- \\
  &  &  Large & \textbf{73.92} & -- & -- && \textbf{74.56} & -- & -- \\
     & &  ECPE-MLL & -- & -- & -- && 66.45 & -- & -- \\
   & &  ECPE-2D & -- & -- & -- && 64.33 & -- & -- \\
   & &  RankCP & -- & -- & -- && 70.21 & -- & -- \\
    % &  MNLI & 29.56 & 94.88 & 62.22 &  &  & \\

%     \midrule
%         \multirow{4}{*}{\rotatebox{90}{\textbf{\small{Fold 2}}}} & \multirow{2}{*}{\rotatebox{90}{\textbf{\small{DD}}}} &  Base & 76.21 & 91.23 & 83.72 & 89.37 & 95.21 & 92.32 \\
%   & &  Large & \textbf{79.52} & \textbf{91.27} & \textbf{85.40} & \textbf{93.05} & \textbf{97.22} & \textbf{95.13} \\
%     \cmidrule{2-9}
%   & \multirow{2}{*}{\rotatebox{90}{\textbf{\small{IEMO}}}} &  Base & 46.12 & \textbf{93.80} & 69.96 & \textbf{65.09} & \textbf{95.60} & \textbf{80.35} \\
%   &  &  Large & \textbf{48.36} & 92.06 & \textbf{70.21} & 61.12 & 95.59 & 78.35 \\
%     \midrule
%         \multirow{4}{*}{\rotatebox{90}{\textbf{\small{Fold 3}}}} & \multirow{2}{*}{\rotatebox{90}{\textbf{\small{DD}}}} &  Base & 74.73 & \textbf{90.33} & \textbf{82.53} & 92.64 & 96.99 & 94.81 \\
%   & &  Large & \textbf{75.79} & 88.43 & 82.11 & \textbf{93.34} & \textbf{97.23} & \textbf{95.29} \\
%     \cmidrule{2-9}
%   & \multirow{2}{*}{\rotatebox{90}{\textbf{\small{IEMO}}}} &  Base & \textbf{51.23} & \textbf{93.70} & \textbf{72.46} & \textbf{63.91} & \textbf{94.55} & \textbf{79.23} \\
%   &  &  Large & 43.00 & 88.47 & 65.74 & 59.03 & 92.21 & 75.62 \\
    \bottomrule
   \end{tabular}}
  \caption{{Results for Causal Emotion Entailment task on the test sets of \RECCONDADD{} and \RECCONDAIE{}. Class-wise F1 score and the overall macro F1 scores are reported. All scores reported at best macro F1 scores. %DD stands for
  % $\xrightarrow{}$ 
 % \RECCONDADD{}, IE for
  % $\xrightarrow{}$ 
  %\RECCONDAIE{}. 
  All models are RoBERTa-based. The cause-pivot emotion extraction setting was used for ECPE-MLL. DD stands for RECCON-DD, IE for RECCON-IE.}}
  \label{tab:cus}
\end{table}

\subsection{Subtask 2: Causal Emotion Entailment}
\label{sec:cus}

The \textit{Causal Emotion Entailment} is a simpler version of the span extraction task. In this task, given a target non-neutral utterance ($U_t$), the goal is to predict which particular utterances in the conversation history $H(U_t)$ are responsible for the non-neutral emotion in the target utterance. Following the earlier setup, we formulate this task with and without historical conversational context.

\subsubsection{Subtask Description}
We consider the following two subtasks:
\paragraph{With Conversational Context (w/ CC)}
We consider the historical conversational context $H(U_t)$ of the target utterance $U_t$ and posit the problem as a triplet classification task: 
% Here 
the tuple $(U_t, U_i, H(U_t))$ is aimed to be classified as positive, $U_i \in C(U_t)$. For negative examples, the tuple $(U_t, U_i, H(U_t))$ should be classified as negative for $U_i \notin C(U_t)$.

\paragraph{Without Conversational Context (w/o CC)}
We posit this problem as a binary sentence pair classification task, where ($U_t$, $U_i$) should be classified as positive as $U_i \in C(U_t)$. For the negative example ($U_t$, $U_i$) where $U_i \notin C(U_t)$, the classification output should be negative.

% Similar to the earlier span extraction task, we term the positive examples as valid instances, and the negative examples as invalid instances.

\subsubsection{Models}
In this paper we consider the following models.
\paragraph{RoBERTa Base and Large}
Similar to subtask 1, we use Transformer-based models to benchmark this task. We use a \code{<CLS>} token and the emotion label ${<}E_t{>}$ of the target utterance $U_t$ in front, and join the pair or triplet elements with \code{<SEP>} in between to create the input. 
% for the models. Finally 
The classification is performed from the corresponding final layer vector of the \code{<CLS>} token. 
% We use the following models: \textbf{RoBERTa Base / Large}: 
We use the \code{roberta-base/-large} models from~\citep{liu2019roberta} as the baselines.
% \paragraph{RoBERTa Large}: The \code{roberta-large} model 
% % from~\citep{liu2019roberta} 
% is used as the other baseline~\citep{liu2019roberta}.
% \paragraph{RoBERTa Fine-tuned on MNLI}: The \code{roberta-large-mnli} model has been fine-tuned on the Multi-NLI dataset~\citep{} and shows impressive performance for sentence pair classification tasks such as natural language inference. We use this model as our other baseline.

\paragraph{ECPE-2D}
\citet{DBLP:conf/acl/DingXY20} proposed an end-to-end approach for emotion cause pair extraction. They use a 2D Transformer network to improve interaction among the utterances.

\paragraph{ECPE-MLL}
\citet{DBLP:conf/emnlp/DingXY20} introduced a joint multi-label approach for emotion cause pair extraction. Specifically, the joint framework comprises two modules: $(i)$ extraction of causal utterances for the target emotion utterance, $(ii)$ extraction of emotion utterance for a causal utterance. Both these modules were trained using a multi-label training scheme. 

\paragraph{RankCP}
\citet{wei-etal-2020-effective} proposed an end-to-end emotion cause pair extraction where first the utterance pairs are ranked and then a one-stage neural approach is applied for inter-utterance correlation modeling that enhances the emotion cause extraction. Specifically, they apply graph attentions to model the interrelations between the
utterances in a dialogue.
ECPE-2D, ECPE-MLL, and RankCP use RoBERTa-base as a sentence encoder in our implementation to facilitate a fair comparison.

\subsubsection{Evaluation Metrics}
We use F1 score for both positive and negative examples, denoted as Pos. F1 and Neg. F1 respectively. We also report the overall macro F1.

\subsection{Results and Discussions}
\label{sec:results}
\cref{tab:cse} shows the 
% experimental 
results of the causal span extraction task where SpanBERT obtains the best performance in both \RECCONDADD{} and \RECCONDAIE{}. SpanBERT outperforms RoBERTa Base in EM$_{Pos}$, and F1$_{Pos}$ metrics. However, the performance of SpanBERT is worse for negative examples, which consequently results in a lower F1 score compared to RoBERTa Base model in both the datasets under ``w/o CC" setting. Contrary to this, the performance of the SpanBERT in the presence of context (w/ CC) is consistently higher than RoBERTa Base with respect to all the metrics in \RECCONDADD{}.
%The performance between SpanBERT and RoBERTa Base is higher when contextual information is not utilized.

In \cref{tab:cus}, we report the performance of the Causal Emotion Entailment task. Under the ``w/o CC'' setting, in Fold $1$, RoBERTa Base outperforms RoBERTa Large by $2$\% in \RECCONDADD{}. In contrast to this, in \RECCONDAIE{}, RoBERTa Large performs better and beats RoBERTa Base by $5.5$\% in Fold $1$. On the other hand, RoBERTa Large outperforms RoBERTa Base in both \RECCONDADD{} and \RECCONDAIE{} under the ``w/ CC'' setting. The performance in \RECCONDAIE{} is consistently worse than in \RECCONDADD{} under various settings in both subtask 1 and 2. We reckon this can be due to multiple reasons mentioned in~\cref{sec:dataset_diffs}, making the task harder on the IEMOCAP split.

%---1)~the dialogues in IEMOCAP are longer, 2)~frequent emotion shifts can be observed across the utterances, and 3)~the existence of long-distant causal spans as opposed to \DailyDialog{}. All these obstacles make the task harder on the IEMOCAP split.

We have also analyzed the performance of the baseline models on the utterances having one or multiple causes. The models consistently perform better for the utterances having only one causal span compared to the ones having multiple causes ($+7$\% on an average calculated over all the settings and models). In the test data of Fold $1$, approximately 38\% of the UCS pairs (which we call as $\overline{\text{Fold 1}}$ ) have their causal spans lie within the target utterances. In \cref{tab:cse} and \ref{tab:cus}, we report the results on $\overline{\text{Fold 1}}$. According to these results, the models perform significantly better on such UCS pairs under all the settings in both the subtasks. % Besides, the models perform much better ($+12$\% on an average calculated over all the settings and models) when the causal span lies in the target utterance $U_t$ instead of its evidence utterance in the historical conversational context. 
The models leverage contextual information for both the subtasks in the ``w/ CC'' setting which substantially improves the performance of the non-contextual (refer to the ``w/o CC'' setting) counterpart. In this setting, SpanBERT obtains the best performance for positive examples in both \RECCONDADD{}, and \RECCONDAIE{}. On the other hand, in the same setting, RoBERTa Large outperforms RoBERTa Base and achieves the best performance in subtask 2.
%Delving deeper, we find that in subtask $1$ and $2$ all the models perform much better in Fold $1$ and $2$ because of the contextual dissimilarity between target utterances and negative examples. 
%The use of context in the models helps to improve the result further in Fold 1 and 2 as it greatly benefits in identifying the contextual discrepancy between target utterances and negative examples.

The low scores of the models in 
% the 
subtasks $1$ and $2$ 
% depict 
show
the difficulty of the tasks. 
% As such, we see 
This implies
% a 
significant room for model improvement in these 
% two 
subtasks of \RECCON{}.
% From the \cref{tab:cus}, we notice 
\cref{tab:cus} shows
that all the complex neural baselines, i.e., ECPE-MLL, ECPE-2D, and RankCP fail to outperform the very simple RoBERTa baselines introduced in this paper. This corroborates the usefulness and importance of these strong baselines, one of the major contributions of this paper.

\section{Further Analysis and Discussion} \label{sec:analysisx}

For further insights into the performance of our models, we analyzed more strategies to create the negative examples: Folds~2 and~3; see \cref{sec:neg} for their description.
% \?{@GELBUKH this para is repeated; explain what Fold 2 and 3 are}

\begin{table}[t!]
  \centering
% \small
  % \resizebox{\linewidth}{!}
  {
  % \tabcolsep=3pt
%\scalebox{0.7}{
   \begin{tabular}{@{}lll@{\hspace{5ex}}cccc@{}c@{\hspace{5ex}}cccc@{}}
    \toprule
   \multicolumn{3}{c}{\multirow3*{\textbf{Model}}} & \multicolumn{4}{c}{\textbf{w/o CC}} && \multicolumn{4}{c}{\textbf{w/ CC}}\\
   \cmidrule{4-7}\cmidrule{9-12}
   % \?{@GELBUKH add empty column}
   & & & EM$_{\text{\it Pos}}$ & F1$_{\text{\it Pos}}$ & F1$_{\text{\it Neg}}$ & F$_1$ && EM$_{\text{\it Pos}}$ & F1$_{\text{\it Pos}}$  & F1$_{\text{\it Neg}}$ & F$_1$ \\
        % Fold 1 -> Fold 1  
       \midrule
   \multirow{4}{*}{\rotatebox{90}{\textbf{\tiny{Fold 1 $\to$ Fold 1}~~}}} 
   & \multirow{2}{*}{\rotatebox{90}{\textbf{{DD}}}} 
   & RoBERTa  & 26.82 & 45.99 & \textbf{84.55} & \textbf{73.82} && 32.63 & 58.17 & 85.85 & 75.45\\
   
  &  & SpanBERT & \textbf{33.26} & \textbf{57.03} & 80.03 & 69.78 && \textbf{34.64} & \textbf{60.00} & \textbf{86.02} & \textbf{75.71} \\
    \cmidrule{2-12}
    & \multirow{2}{*}{\rotatebox{90}{\textbf{{IE}}}} & RoBERTa  & \09.81 & 18.59 & \textbf{93.45} & \textbf{87.60} && 10.19 & 26.88 & \textbf{91.68} & \textbf{84.52}\\
    
  &  & SpanBERT & \textbf{16.20} & \textbf{30.22} & 87.15 & 77.45 && \textbf{22.41}  & \textbf{37.80} & 90.54 & 82.86 \\
       % Fold 1 -> Fold 2
  
      \midrule
       \multirow{4}{*}{\rotatebox{90}{\textbf{\tiny{Fold 1 $\to$ Fold 2}~~}}} & \multirow{2}{*}{\rotatebox{90}{\textbf{{DD}}}} & RoBERTa  & 26.82 & 45.99 & 83.52 & 72.66 && \textbf{32.95} & \textbf{59.02} & \textbf{95.36} & \textbf{87.63} \\
    &  & SpanBERT & \textbf{33.26} & \textbf{57.03} & \textbf{84.02} & \textbf{74.80} && 32.37 & 57.04 & 95.01 & 87.00 \\
    \cmidrule{2-12}
    & \multirow{2}{*}{\rotatebox{90}{\textbf{{IE}}}} & RoBERTa  & \09.81 & 18.59 & \textbf{92.18} & \textbf{85.41}  && 10.93 & 28.26 & 95.49 & 90.85 \\
  &  & SpanBERT & \textbf{16.20} & \textbf{30.22} & 88.63 & 79.80 && \textbf{24.07} & \textbf{40.57} & \textbf{96.28} & \textbf{92.41} \\
      % Fold 1 -> Fold 3
      \midrule
       \multirow{4}{*}{\rotatebox{90}{\textbf{\tiny{Fold 1 $\to$ Fold 3}~~}}} & \multirow{2}{*}{\rotatebox{90}{\textbf{{DD}}}} & RoBERTa  & 26.82 & 45.99 & \textbf{81.50} & \textbf{70.26} && \textbf{32.95} & \textbf{59.02} & \textbf{95.37} & \textbf{87.65} \\
  &  & SpanBERT & \textbf{33.26} & \textbf{57.03} & 79.65 & 69.83 && 32.31 & 56.99 & 94.92 & 86.87 \\
    \cmidrule{2-12}
    & \multirow{2}{*}{\rotatebox{90}{\textbf{{IE}}}} & RoBERTa  & \09.81 & 18.59 & \textbf{91.82} & \textbf{84.83} && 10.93 & 28.26 & 95.47 & 90.81 \\
  &  & SpanBERT & \textbf{16.20} & \textbf{30.22} & 86.95 & 77.25 && \textbf{24.07} & \textbf{40.57} & \textbf{96.28}  & \textbf{92.41}  \\
   %  Fold 2
    \midrule
       \multirow{4}{*}{\rotatebox{90}{\textbf{\tiny{Fold 2 $\to$ Fold 2}~~}}} & \multirow{2}{*}{\rotatebox{90}{\textbf{{DD}}}} & RoBERTa  & \textbf{33.26} & 58.44 & 90.14 & 82.19 && 41.61 & 73.57 & \textbf{99.98} & 92.04 \\
  &  & SpanBERT & 32.31 & \textbf{58.61} & \textbf{90.20} & \textbf{82.29} && \textbf{41.97} & \textbf{74.85} & 99.94 & \textbf{92.43} \\
    \cmidrule{2-12}
    & \multirow{2}{*}{\rotatebox{90}{\textbf{{IE}}}} & RoBERTa  & 15.93 & 31.74 & \textbf{92.93} & \textbf{86.50} && 30.28 & 59.14 & \textbf{99.43} & 94.58 \\
  &  & SpanBERT & \textbf{22.13} & \textbf{38.84} & 90.37 & 82.49 && \textbf{32.50} & \textbf{65.45} & 98.37 & \textbf{95.50} \\
  
  % Fold 2 -> Fold 1
  
      \midrule
       \multirow{4}{*}{\rotatebox{90}{\textbf{\tiny{Fold 2 $\to$ Fold 1}~~}}} & \multirow{2}{*}{\rotatebox{90}{\textbf{{DD}}}} & RoBERTa  & \textbf{33.26} & 58.44 & 71.29 & 60.45 && \textbf{36.06} & \textbf{65.04} & \00.19 & \textbf{17.12} \\
    &  & SpanBERT & 32.31 & \textbf{58.61} & \textbf{72.52} & \textbf{61.70} && 31.52 & 60.81 & \textbf{\00.67} & 16.19 \\
    \cmidrule{2-12}
    & \multirow{2}{*}{\rotatebox{90}{\textbf{{IE}}}} & RoBERTa  & 15.93 & 31.74 & \textbf{90.70} & \textbf{82.91} && \textbf{22.96} & 46.87 & \04.66 & \06.35 \\
  &  & SpanBERT & \textbf{22.13} & \textbf{38.84} & 85.03 & 74.34 && 21.85 & \textbf{49.18} & \textbf{\06.36} & \textbf{\07.40} \\
  
    %  Fold 3
    \midrule
       \multirow{4}{*}{\rotatebox{90}{\textbf{\tiny{Fold 3 $\to$ Fold 3}~~}}} & \multirow{2}{*}{\rotatebox{90}{\textbf{{DD}}}} & RoBERTa  & 28.72 & 51.32 & \textbf{90.06} & \textbf{82.11} && 41.29 & 74.95 & \textbf{99.94} & 92.44 \\
  &  & SpanBERT & \textbf{30.62} & \textbf{54.96} & 89.41 & 81.21 && \textbf{42.61} & \textbf{75.36} & 99.93 & \textbf{92.46} \\
    \cmidrule{2-12}
    & \multirow{2}{*}{\rotatebox{90}{\textbf{{IE}}}} & RoBERTa  & 14.54 & 26.51 & \textbf{93.68} & \textbf{87.79} && 24.35 & 53.46 & 97.84 & 94.08 \\
  &  & SpanBERT & \textbf{17.41} & \textbf{31.75} & 91.85 & 84.86 && \textbf{32.87} & \textbf{62.70} & \textbf{99.54} & \textbf{95.11} \\
  
  % Fold 3 -> Fold 1
      \midrule
       \multirow{4}{*}{\rotatebox{90}{\textbf{\tiny{Fold 3 $\to$ Fold 1}~~}}} & \multirow{2}{*}{\rotatebox{90}{\textbf{{DD}}}} & RoBERTa  & 28.72 & 51.32 & \textbf{75.55} & 64.31 && \textbf{37.22} & \textbf{69.64} & \textbf{\00.90} & \textbf{18.59} \\
  &  & SpanBERT & \textbf{30.62} & \textbf{54.96} & 75.49 & \textbf{64.46} && 31.94 & 60.81 & \00.15 & 16.00\\
    \cmidrule{2-12}
    & \multirow{2}{*}{\rotatebox{90}{\textbf{{IE}}}} & RoBERTa  & 14.54 & 26.51 & \textbf{92.33} & \textbf{85.61} && 21.20 & \textbf{48.34} & \textbf{11.42} & \textbf{\09.76}\\
  &  & SpanBERT & \textbf{17.41} & \textbf{31.75} & 89.41 & 80.94 && \textbf{21.48} & 45.49 & \04.01 & \05.84 \\
    \bottomrule
   \end{tabular}
  }
%   }
  \caption{{Results for Causal Span Extraction task on the test sets of \RECCONDADD{} and \RECCONDAIE{}. All scores are in percentage and are reported at best validation F1 scores. RoBERTa stands for RoBERTa Base, DD for \RECCONDADD{}, IE for \RECCONDAIE{}. Fold $i$ $\to$ Fold $j$ means trained on Fold $i$, tested on Fold $j$.}}
  \label{tab:cse2x}
\end{table}

The use of context (w/ CC) in the baseline models improves the results (see Tables~\ref{tab:cse2x} and~\ref{tab:cus2x}) in Folds~2 and~3 as it highlights the contextual discrepancy or coherence between the target utterance and context which should strongly aid in identifying randomly generated negative samples from the rest. For the positive examples, we achieve a much better score in Folds~2 and~3 as compared with Fold~1 (see Tables~\ref{tab:cse} and~\ref{tab:cus}) for both ``w/o CC" and ``w/ CC" constraints. However, this does not validate Folds~2 and~3 as better training datasets than Fold~1. We confirm this by training the models on Folds~2 and~3 and evaluating them on Fold~1. These two experiments are denoted as {Fold~2 $\to$ Fold~1} and {Fold~3 $\to$ Fold~1}, respectively, and the corresponding results are reported in Tables~\ref{tab:cse2x} and~\ref{tab:cus2x}. The outcomes of these experiments, as shown in Tables~\ref{tab:cse2x} and~\ref{tab:cus2x}, show abysmal performance by the baseline models on the negative examples in Fold~1.

This may be ascribed to the fundamental difference between Fold~1 and Folds~2 and~3. Negative samples in Folds~2 and~3 are easily identifiable, as compared to Fold~1, as all the model needs to do to judge the absence of a causal span in the context is to detect the contextual incoherence of the target utterance with the context. Models fine-tuned on BERT and SpanBERT are expected to perform well at deciding contextual incoherence. Identifying negative samples in Fold~1, however, requires more sophisticated and non-trivial approach as the target utterances are, just as the positive examples, contextually coherent with the context. As such, a model that correlates contextual incoherence with negative samples naturally performs poorly on Fold~1.

% This may be attributed to the models' ability to learn to measure the similarity between target utterance with causal utterance and contextual history implicitly while solving the subtasks. This consequently boosts the baseline models' ability to discriminate relevant context and causal utterance from irrelevant ones\?{check} with respect to the target utterance. In Fold 2 and 3, the discrimination is easier as the negative examples are created by collating utterances from other dialogues that are irrelevant to the target utterance. Contrary to this, negative examples in Fold 1 are created using the historical utterances of a target utterance. Hence, when we trained the models in Fold 2 and 3 and evaluate them in Fold 1, they suffer from the confusion that surfaced due to the strong relevance between contextual history and target utterance in the negative examples which as a result reduces the performance.
The $F1_{Neg}$ scores for {Fold~2 $\to$ Fold~1}, and {Fold~3 $\to$ Fold~1} modes under both ``w/o CC" and ``w/ CC" settings are adversely affected by the low precision of the models in both the subtasks. In other words, the baseline models in these two modes perform poor in extracting empty spans from the ground truth negative examples in subtask~1 and also classify most of the negative examples as positive in subtask~2.
\begin{table}[t!]
  \centering
%  \small
 % \resizebox{\linewidth}{!}
 {
  % \tabcolsep=3pt
%\scalebox{0.6}{
   \begin{tabular}{@{}lll@{\hspace{7ex}}ccc@{}c@{\hspace{6ex}}ccc@{}}
    \toprule
      \multicolumn{3}{c}{\multirow3*{\textbf{Model}}} & \multicolumn{3}{c}{\textbf{w/o CC}} && \multicolumn{3}{c}{\textbf{w/ CC}}\\
      \cmidrule{4-6}\cmidrule{8-10}
    && & Pos. F1 & Neg. F1 & macro F1 && Pos. F1 & Neg. F1 & macro F1\\

%    \midrule
%     \multirow{4}{*}{\rotatebox{90}{\textbf{\small{Fold 1}}}} & \multirow{2}{*}{\rotatebox{90}{\textbf{\small{DD}}}} &  Base & \textbf{56.64} & 85.13 & \textbf{70.88} & 64.28 & \textbf{88.74} & 76.51 \\
%   & &  Large & 50.48 & \textbf{87.35} & 68.91 & \textbf{66.23} & 87.89 & \textbf{77.06} \\
%     % &  MNLI & 55.19 & 87.95 & 71.57 & & & \\
%     \cmidrule{2-9}
%   & \multirow{2}{*}{\rotatebox{90}{\textbf{\small{IEMO}}}} &  Base & 25.98 & 90.73 & 58.36 & 28.02 & \textbf{95.67} & 61.85\\
%   &  &  Large & \textbf{32.34} & \textbf{95.61} & \textbf{63.97} & \textbf{40.83} & \textbf{95.68} & \textbf{68.26} \\
%     % &  MNLI & 29.56 & 94.88 & 62.22 &  &  & \\
% Fold 1
    \midrule
    \multirow{4}{*}{\rotatebox{90}{\textbf{\tiny{Fold 1 $\to$ Fold 1}~~}}} & \multirow{2}{*}{\rotatebox{90}{\textbf{{DD}}}} &  Base & \textbf{56.64} & 85.13 & \textbf{70.88} && 64.28 & \textbf{88.74} & 76.51 \\
   & &  Large & 50.48 & \textbf{87.35} & 68.91 && \textbf{66.23} & 87.89 & \textbf{77.06} \\
    % &  MNLI & 55.19 & 87.95 & 71.57 && & & \\
    \cmidrule{2-10}
   & \multirow{2}{*}{\rotatebox{90}{\textbf{{IE}}}} &  Base & 25.98 & 90.73 & 58.36 && 28.02 & \textbf{95.67} & 61.85\\
  &  &  Large & \textbf{32.34} & \textbf{95.61} & \textbf{63.97} && \textbf{40.83} & \textbf{95.68} & \textbf{68.26} \\
    % Fold 1 -> Fold 2
        \midrule
        \multirow{4}{*}{\rotatebox{90}{\textbf{\tiny{Fold 1 $\to$ Fold 2}~~}}} & \multirow{2}{*}{\rotatebox{90}{\textbf{{DD}}}} &  Base & \textbf{57.50}  & 82.71  & 70.11  && 59.06  & 86.91  & 72.98 \\
   & &  Large & 56.13  & \textbf{88.33}  & \textbf{72.23}  && \textbf{60.09} & \textbf{88.00} & \textbf{74.04} \\
    \cmidrule{2-10}
   & \multirow{2}{*}{\rotatebox{90}{\textbf{{IE}}}} &  Base & 32.60  & 89.99  & 61.30  && 27.14  & 94.16  & 60.65 \\
  &  &  Large & \textbf{36.61}  & \textbf{94.60} & \textbf{65.60}  && \textbf{37.59}  & \textbf{94.63}  & \textbf{66.11} \\  
    % Fold 1 -> Fold 3
        \midrule
    \multirow{4}{*}{\rotatebox{90}{\textbf{\tiny{Fold 1 $\to$ Fold 3}~~}}} & \multirow{2}{*}{\rotatebox{90}{\textbf{{DD}}}} &  Base  & 57.52 & 82.72  & 70.12  && 49.30  & 79.27  & 64.29 \\
   & &  Large & \textbf{56.04}  & \textbf{88.28}  & \textbf{72.16}  && \textbf{60.63}  & \textbf{88.30}  & \textbf{74.46} \\
    \cmidrule{2-10}
   & \multirow{2}{*}{\rotatebox{90}{\textbf{{IE}}}} &  Base & 33.24  & 90.30  & 61.77  && 23.83  & 92.97  & 58.40 \\
  &  &  Large & \textbf{36.55}  & \textbf{94.59}  & \textbf{65.57}  && \textbf{37.87}  & \textbf{94.69}  & \textbf{66.28} \\  
    % Fold 2 
     \midrule
        \multirow{4}{*}{\rotatebox{90}{\textbf{\tiny{Fold 2 $\to$ Fold 2}~~}}} & \multirow{2}{*}{\rotatebox{90}{\textbf{{DD}}}} &  Base & 76.21 & 91.23 & 83.72 && 89.37 & 95.21 & 92.32 \\
   & &  Large & \textbf{79.52} & \textbf{91.27} & \textbf{85.40} && \textbf{93.05} & \textbf{97.22} & \textbf{95.13} \\
    \cmidrule{2-10}
   & \multirow{2}{*}{\rotatebox{90}{\textbf{{IE}}}} &  Base & 46.12 & \textbf{93.80} & 69.96 && \textbf{65.09} & \textbf{95.60} & \textbf{80.35} \\
  &  &  Large & \textbf{48.36} & 92.06 & \textbf{70.21} && 61.12 & 95.59 & 78.35 \\

  % Fold 2 -> Fold 1
      \midrule
        \multirow{4}{*}{\rotatebox{90}{\textbf{\tiny{Fold 2 $\to$ Fold 1}~~}}} & \multirow{2}{*}{\rotatebox{90}{\textbf{{DD}}}} &  Base  & \textbf{52.52} & \textbf{75.51}  & \textbf{64.02}  && 41.86 & \03.25  & 22.55 \\
   & &  Large & 51.57 & 67.58  & 59.57  && \textbf{43.25}  & \textbf{19.95} & \textbf{31.60} \\
    \cmidrule{2-10}
   & \multirow{2}{*}{\rotatebox{90}{\textbf{{IE}}}} &  Base & \textbf{31.51}  & \textbf{92.09}  & \textbf{61.80}  && 25.22  & 74.69  & 49.96 \\
  &  &  Large & 29.64  & 87.68  & 58.66  && \textbf{26.30} & \textbf{76.44}  & \textbf{51.37} \\
  
   % Fold 3
    \midrule
        \multirow{4}{*}{\rotatebox{90}{\textbf{\tiny{Fold 3 $\to$ Fold 3}~~}}} & \multirow{2}{*}{\rotatebox{90}{\textbf{{DD}}}} &  Base & 74.73 & \textbf{90.33} & \textbf{82.53} && 92.64 & 96.99 & 94.81 \\
   & &  Large & \textbf{75.79} & 88.43 & 82.11 && \textbf{93.34} & \textbf{97.23} & \textbf{95.29} \\
    \cmidrule{2-10}
   & \multirow{2}{*}{\rotatebox{90}{\textbf{{IE}}}} &  Base & \textbf{51.23} & \textbf{93.70} & \textbf{72.46} && \textbf{63.91} & \textbf{94.55} & \textbf{79.23} \\
  &  &  Large & 43.00 & 88.47 & 65.74 && 59.03 & 92.21 & 75.62 \\
  
  % Fold 3 -> Fold 1
      \midrule
        \multirow{4}{*}{\rotatebox{90}{\textbf{\tiny{Fold 3 $\to$ Fold 1}~~}}} & \multirow{2}{*}{\rotatebox{90}{\textbf{{DD}}}} &  Base  & \textbf{52.02} & \textbf{74.59}  & \textbf{63.31}  && 41.64 & \02.99  & 22.31 \\
   & &  Large & 51.53  & 65.76  & 58.65  && \textbf{41.86} & \textbf{\04.89} & \textbf{23.38} \\
    \cmidrule{2-10}
   & \multirow{2}{*}{\rotatebox{90}{\textbf{{IE}}}} &  Base & \textbf{34.74}  & \textbf{91.46}  & \textbf{63.10}  && \textbf{19.13}  & \textbf{54.25}  & \textbf{36.69} \\
  &  &  Large & 27.58  & 84.13  & 55.86  && 18.33 &  48.01  & 33.17 \\
    \bottomrule
   \end{tabular}
  }
  \caption{{Results for Causal Emotion Entailment task on the test sets of \RECCONDADD{} and \RECCONDAIE{}. Class wise F1 scores and the overall macro F1 scores are reported. All scores reported at best macro F1 scores. DD stands for \RECCONDADD{}, IE for \RECCONDAIE{}. All models are RoBERTa-based models. Fold $i$ $\to$ Fold $j$ means trained on Fold $i$, tested on Fold $j$.}}
  \label{tab:cus2x}
\end{table}

On the other hand, we do not observe any significant performance drop for either negative or positive examples when the models trained in Fold 1 are evaluated in Folds~2 and~3. This affirms the superiority of Fold~1 as a training dataset. Besides, note that Fold~1 is a more challenging and practical choice than the rest of the two folds as in real scenarios, we need to identify causes of emotions within a single dialogue by reasoning over the utterances in it.

\section{Challenges of the Task} 
\label{sec:challengesx}

This section identifies several examples that indicate the need for \textbf{complex reasoning} to solve the causal span extraction task. Abilities to accurately reason will help validate if a candidate span is causally linked to the target emotion. We believe these pointers would help further research on this dataset and solving the task in general.

\begin{figure}[t!]
    \centering
    \includegraphics[width=60ex]{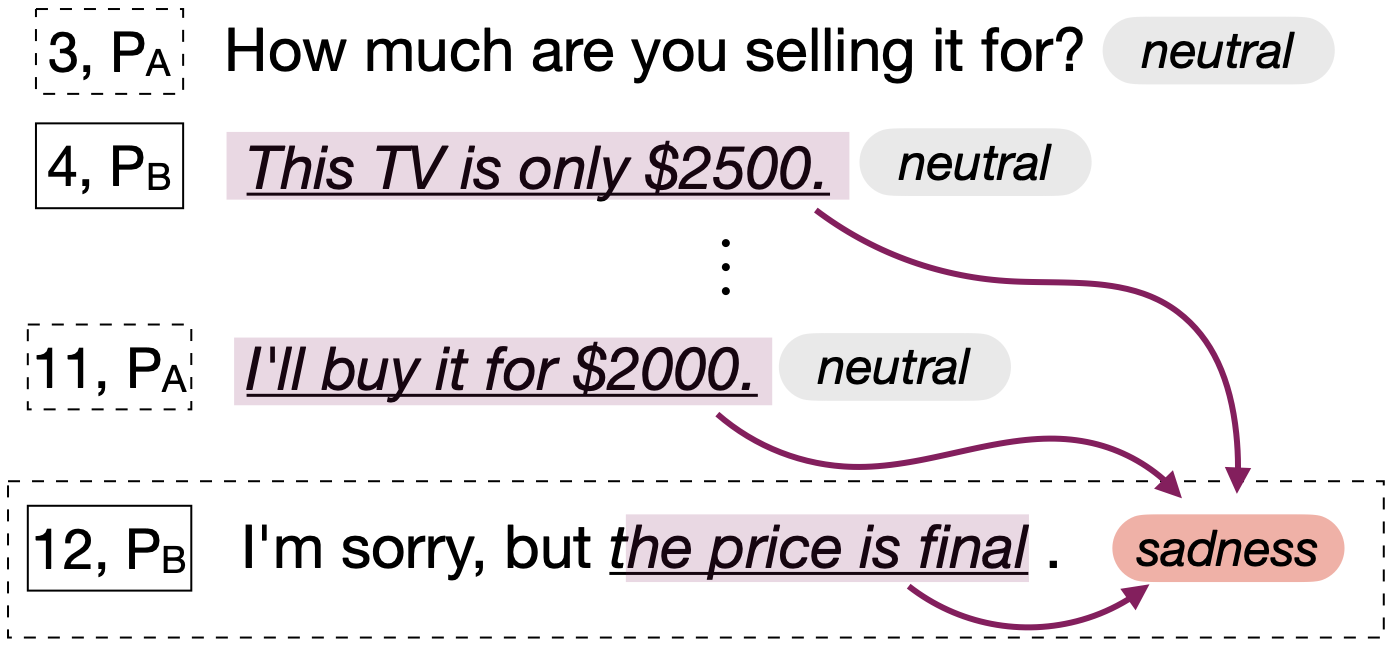}
    \caption{{In this example, $P_B$, in utterance $12$, is sad because of failing to negotiate the desired amount to sell a TV. While ``\textit{the price is final}" is a valid causal span, one also needs to identify the discussion where $P_A$ is ready to pay only \$2000, which is significantly lesser than the originally quoted \$2500.}}
    \label{fig:numerical_reasoning}
\end{figure}

\paragraph{Amount of Spans} 
%\hl{This should be mentioned only if multi-span extraction is done.}
One of the primary challenges of this task is determining the set of spans that can sufficiently be treated as the cause for a target emotion. The spans should have coverage to be able to formulate logical reasoning steps (performed implicitly by annotators) that include skills such as numerical reasoning (see \cref{fig:numerical_reasoning}), among others.

\paragraph{Emotional Dynamics} Understanding emotional dynamics in conversations is closely tied with emotion cause identification. As shown in our previous sections, many causal phrases in the dataset depend on the inter-personal event/concept mentions, emotions, and self-influences (sharing causes). We also observe that emotion causes may be present across multiple turns, thus requiring the ability to model long-term information. Emotions of the contextual utterances help in this modeling. In fact, without the emotional information of the contextual utterances, our annotators found it difficult to annotate emotion causes in the dataset. Understanding cordial greetings, conflicts, agreements, and empathy are some of the many scenarios where contextual emotional dynamics play a significant role. 

% \subsubsection{Conversational Discourse} In addition to emotional dynamics across turns, we observe that emotional cause is also related to discourse in a conversation. Latent causes in the utterances can sometimes be explained with dialogue acts e.g.,

% \begin{exe}
% \ex \textit{A (happy)}: Hello! Welcome to our store, Madam!\\
% \textit{B (neutral)}: Thanks. I need a luxury watch.
% \label{ex:discourse}
% \end{exe}

% In example \ref{ex:discourse}, the cause of utterance 1 is latent but the dialogue act label \emph{greeting} can define it well.

\paragraph{Commonsense Knowledge} Extracting emotion causes in conversations comprises complex reasoning steps, and commonsense knowledge is an integral part of this process. The role of commonsense reasoning in emotion cause recognition is more evident when the underlying emotion the cause is latent. Consider the example below:
\begin{exe}
\ex {$P_A$ (\emo{happy})}: \textit{Hello, thanks for calling 123 Tech Help, I'm Todd. How can I help you?}\\
{$P_B$ (\emo{fear})}: \textit{Hello ? Can you help me ? My computer ! Oh man ...}
\label{ex:latent}
\end{exe}
In this case, $P_A$ is happily offering help to $P_B$. The cause of happiness in this example is due to the event ``\textit{greeting}" or intention to offer help. On the other hand, $P_B$ is fearful because of his/her \textit{broken computer}. The causes of elicited emotions by both the speakers can only be inferred using commonsense knowledge.

\begin{figure}[t!]
    \centering
    \includegraphics[width=60ex]{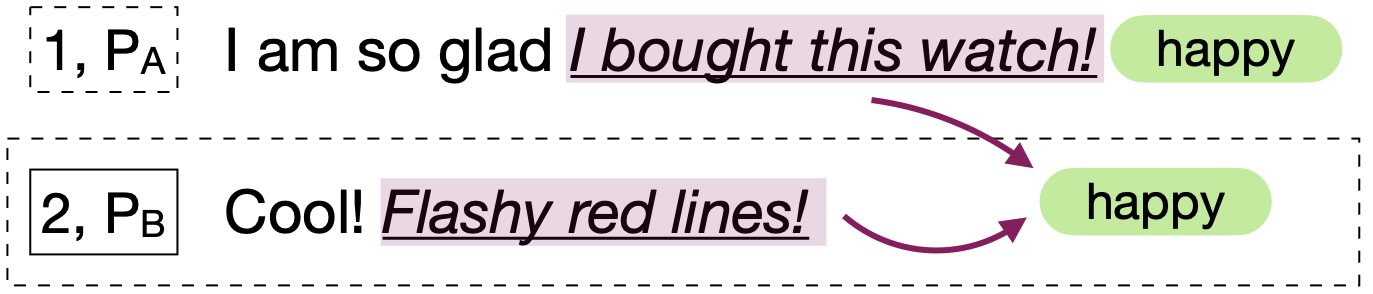}
    \caption{{In this example, the emotion cause for utterance 2 may lie in phrases spoken by (and for) the counterpart ($P_A$) and not the target speaker ($P_B$) i.e., ``\textit{flashy red lines}'' in $P_B$'s utterance points to the property of the ``\textit{watch}'' that $P_A$ bought. One needs to infer such co-referential links to extract the correct causal spans.}}
    \label{fig:pronoun_mismatch}
\end{figure}

\paragraph{Complex Co-Reference} While in narratives, co-references are accurately used and often explicit, it is not the case in dialogues (see~\cref{fig:pronoun_mismatch}).  

\paragraph{Exact vs. Perceived Cause} At times, the complex and informal nature of conversations prohibits the extraction of exact causes. In such cases, our annotators extract the spans that can be perceived as the respective cause. These causal spans can be rephrased to represent the exact cause for the expressed emotion. For example,

\begin{exe}
\ex {$P_A$ (\emo{neutral})}: \textit{How can I help you Sir?.}\\
{$P_B$ (\emo{frustrated})}: \textit{I just want my flip phone to work----that's all I need.}
\label{ex:exact1}
% \ex \textit{A (sadness)}: You look very sad.\\
% \textit{B (sadness)}: I wish I was present there to support him.\\
% \label{ex:exact2}
\end{exe}
%
% In example \ref{ex:exact1} and \ref{ex:exact2}, the causes lie in the following sentences---``\textit{I just want my flip phone to work}", and ``\textit{I wish I was present there to support him}". However the exact causes are the rephrased versions of these spans---``\textit{My flip phone is not working}", and ``\textit{I was not present there to support him}". Special dialogue act labels such as \emph{goal achieved} and \emph{goal not-achieved} can also be adopted to describe the causes of this type.
%
In this example, the cause lies in the sentence ``\textit{I just want my flip phone to work}", with the exact cause meaning of ``\textit{My flip phone is not working}". Special dialogue-act labels such as \emph{goal achieved} and \emph{goal not-achieved} can also be adopted to describe such causes.

\paragraph{From Cause Extraction to Causal Reasoning}

\begin{figure}[t!]
    \centering
    \includegraphics[width=60ex]{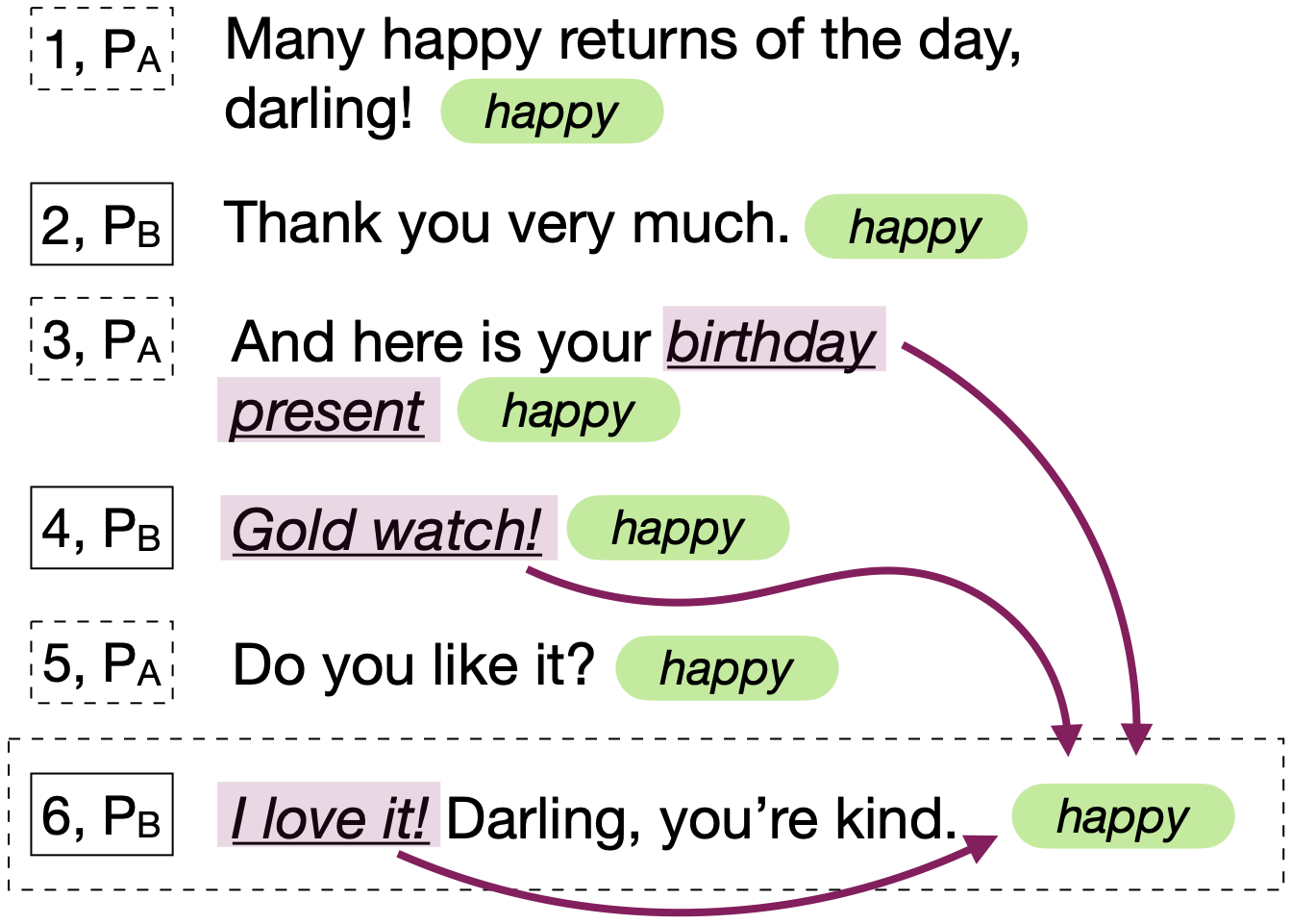}
    \caption{{In this example, the cause for the happy state of $P_B$ (utterance 6) is corroborated by three indicated spans. First, $p_B$ gets happy over receiving a ``\textit{birthday present}" (utterance 3) which is a ``\textit{gold watch}" (utterance 4). Then, the emotion evoked by the 4$^{th}$ utterance is propagated into $P_B$'s next utterance where it is confirmed that $P_B$ loves the gift (``\textit{I love it!}"). Performing temporal reasoning over these three spans helps understand that $P_B$ is happy because of liking a present received as a birthday gift.}}
    \label{fig:temporal_reasoning}
\end{figure}

\begin{figure}[ht!]
    \centering
    \includegraphics[width=60ex]{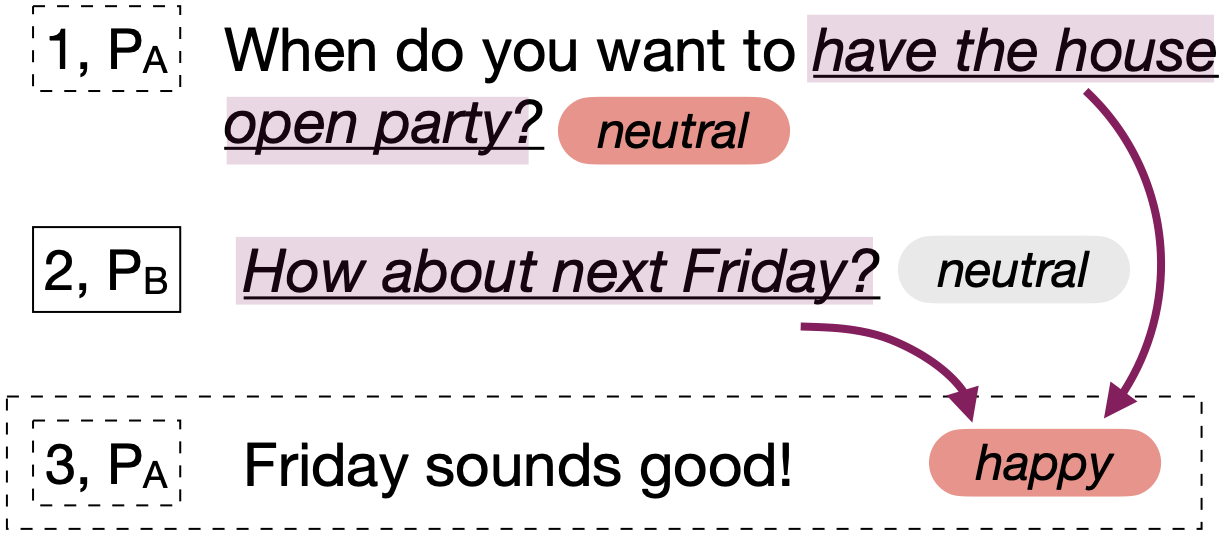}
    \caption{{An example of emotional reasoning where the \emph{happiness} in utterance~3 is caused by the triggers in utterances~1 and~2.}}
    \label{fig:csk_exx}
\end{figure}

Extracting causes of utterances involve reasoning steps. In this work, we do not ask our annotators to explain the reasoning steps pertaining to the extracted causes. However, one can still sort the extracted causes of an utterance according to their temporal order of occurrence in the dialogue. The resulting sequence of causes can be treated as a participating subset of the reasoning process as shown in \cref{fig:temporal_reasoning}. In the future, this dataset can be extended by including reasoning procedures. However, coming up with an optimal set of instructions for the annotators to code the reasoning steps is one of the major obstacles. \cref{fig:csk_exx} also demonstrates the process of reasoning where utterance $1$ and $2$ are the triggers of \emo{happy} emotion in the utterance $3$. However, the reasoning steps that are involved to extract these causes can be defined as: $P_A$ is happy because his/her goal to participate in the \textit{house open party} is achieved after the confirmation of $P_B$ who will organize the \textit{house open party}. This reasoning includes understanding discourse~\cite{chakrabarty-etal-2019-ampersand}, logic and leveraging commonsense knowledge.

More generally, 
% \textbf{\underline{e}motion \underline{c}ausal \underline{r}easoning \underline{i}n \underline{c}on\-ver\-sa\-tions} 
\textbf{emotion causal reasoning in conversations}
extends the task of identifying emotion cause to determining the \textbf{function} and \textbf{explanation} of why the stimuli or triggers evoke the emotion in the target utterance. 
\textbf{Evidence utterance ($U^t_c$):} An utterance containing a span that is the target utterance's emotion cause. As there can be multiple evidence utterances of $U_t$, we represent the set all evidence utterances as $C_{U_t} = \{U^t_c\ \mid c \leq t\}$ and $C_{U_t} \subseteq H_{U_t}$. 

% \cref{fig:example} presents another example of such reasoning process, where~$P_B$'s \emo{frustration} in utterance~$2$ is caused by the unexpected question in utterance~$1$ by $P_A$ with emotion \emo{anger}. In utterance~$3$, $P_A$'s emotion shifts from \emo{anger} to \emo{frustration} as he or she further elaborates the assumptions behind the utterance~$1$. The emotion \emo{frustration} continues until utterance~$13$, where $P_A$ expresses \emo{sadness} while sharing personal experience. 
% We will show two examples of emotion causal reasoning in \cref{fig:temporal_reasoning} and~\ref{fig:numerical_reasoning} below.

% \begin{figure}[t] 
%     \centering 
%     \includegraphics[width=0.9\linewidth]{emotion_understanding.png} 
%     \caption{{Emotion reasoning in conversations.}}
%     \label{fig:example}
% \end{figure}

% \section{Discussions}
% \hl{Show performance comparison on utterances with one vs multiple cause}
% \hl{Show performance comparison on utterances whose cause lie in the context vs in the same utterance.}

%%%%%%%%%%%%%%%%%%%%%%%%%%%%%%%%%%%%%%%%%
%\section{From Cause Extraction to Reasoning}
%%%%%%%%%%%%%%%%%%%%%%%%%%%%%%%%%%%%%%%%%

%\hl{to complete}
\section{Connection to Interpretability of the Contextual Models}
One of the advantages of identifying the causes of emotions in conversations is its role in interpreting a model's predictions. We reckon two situations where emotion cause identification can be useful to verify the interpretability of the contextual emotion recognition models that rely on attention mechanisms to count on the context:

\begin{itemize}[leftmargin=*]
    \item In conversations, utterances may not contain any explicit emotion bearing words or sound neutral on the surface but still carry emotions that can only be inferred from the context. In these cases, one can probe contextual models by dropping the causal utterances that contribute significantly to evoke emotion in the target utterance. It would be interesting to observe whether the family of deep networks that rely on attention mechanisms for context modeling e.g., transformer assign higher probability scores to causal contextual utterances in order to make correct predictions.
    \item As discussed in Section 5, the cause can be present in the target utterance and the model may not need to cater contextual information to predict the emotion. In such cases, it would be worth checking whether attention-based models assign high probability scores to the spans in the target utterance that contribute to the causes of its emotion. 
\end{itemize}

One should also note that a model does not always need to identify the cause of emotions to make correct predictions. For example, 
\begin{exe}
\ex {$P_A$ (\emo{happy})}: \textit{Germany won the match!}\\
{$P_B$ (\emo{happy})}: \textit{That's great!}
\end{exe}
Here, a model can predict the emotion of $P_B$ by just leveraging the cues present in the corresponding utterance. However, the utterance by $P_B$ is just an expression and the cause of the emotion is an event ``\textit{Germany won the match}". Nonetheless, identifying the causes of emotions expressed in a conversation makes the model trustworthy, interpretable, and explainable. 

\section{Conclusion}
%%%%%%%%%%%%%%%%%%%%%%%%%%%%%%%%%%%%%%%%%
% In this work, 
We have addressed the problem of \textbf{R}ecognizing \textbf{E}motion \textbf{C}ause in \textbf{CON}versations and introduced a new dialogue-level dataset, \RECCONDA{}, 
% It is a dialogue-level dataset 
containing more than 1,126 dialogues (dyadic conversations) and 10,600 utterance causal span pairs. We identified various emotion types and key challenges that make the task 
% of \RECCON{} 
extremely challenging. Further, we 
% also 
proposed two subtasks and formulated Transformer-based strong baselines to address these subtasks.
% Our 
% % proposed 
% dataset only incorporates dyadic conversations. 

Future work will target the analysis of emotion cause in multi-party settings. We also plan to annotate the reasoning steps involved in identifying causal spans of elicited emotions in conversations.
Another direction of future work is to extend the approach to multi-modal setting, both in terms of transferring our annotation to the multi-modal data where such data are available (the part of our dataset extracted from IEMOCAP) and in terms of the benchmark algorithms.

%\begin{acknowledgements}
%\?{acks}
%\end{acknowledgements}

\section*{Conflict of interest}
The authors declare that they have no conflict of interest.
\section*{Compliance with Ethical Standards}
\begin{itemize}
    % \item The authors did not receive support from any organization for the submitted work.
    \item This article does not contain any studies with human participants or animals performed by any of the authors.
    \item All authors certify that they have no affiliations with or involvement in any organization or entity with any financial interest or non-financial interest in the subject matter or materials discussed in this manuscript.
\end{itemize}

\bibliographystyle{spbasic}      % basic style, author-year citations
\bibliography{refs}

\begin{thebibliography}{38}
\providecommand{\natexlab}[1]{#1}
\providecommand{\url}[1]{{#1}}
\providecommand{\urlprefix}{URL }
\expandafter\ifx\csname urlstyle\endcsname\relax
  \providecommand{\doi}[1]{DOI~\discretionary{}{}{}#1}\else
  \providecommand{\doi}{DOI~\discretionary{}{}{}\begingroup
  \urlstyle{rm}\Url}\fi
\providecommand{\eprint}[2][]{\url{#2}}

\bibitem[{Ameer et~al.(2020)Ameer, Ashraf, Sidorov, and Adorno}]{Iqra}
Ameer I, Ashraf N, Sidorov G, Adorno HG (2020) Multi-label emotion
  classification using content-based features in {Twitter}. Computaci\'on y
  Sistemas 24(3):1159--1164, \doi{10.13053/CyS-24-3-3476}

\bibitem[{Brandsen et~al.(2020)Brandsen, Verberne, Wansleeben, and
  Lambers}]{brandsen-etal-2020-creating}
Brandsen A, Verberne S, Wansleeben M, Lambers K (2020) Creating a dataset for
  named entity recognition in the archaeology domain. In: Proceedings of the
  12th Language Resources and Evaluation Conference, European Language
  Resources Association, Marseille, France, pp 4573--4577,
  \urlprefix\url{https://www.aclweb.org/anthology/2020.lrec-1.562}

\bibitem[{Busso et~al.(2008)Busso, Bulut, Lee, Kazemzadeh, Mower, Kim, Chang,
  Lee, and Narayanan}]{iemocap}
Busso C, Bulut M, Lee CC, Kazemzadeh A, Mower E, Kim S, Chang JN, Lee S,
  Narayanan SS (2008) {IEMOCAP: Interactive emotional dyadic motion capture
  database}. Language Resources and Evaluation 42(4):335--359

\bibitem[{Chakrabarty et~al.(2019)Chakrabarty, Hidey, Muresan, McKeown, and
  Hwang}]{chakrabarty-etal-2019-ampersand}
Chakrabarty T, Hidey C, Muresan S, McKeown K, Hwang A (2019) {AMPERSAND}:
  Argument mining for {PERS}u{A}sive o{N}line discussions. In: Proceedings of
  the 2019 Conference on Empirical Methods in Natural Language Processing and
  the 9th International Joint Conference on Natural Language Processing
  (EMNLP-IJCNLP), Association for Computational Linguistics, Hong Kong, China,
  pp 2933--2943, \doi{10.18653/v1/D19-1291},
  \urlprefix\url{https://www.aclweb.org/anthology/D19-1291}

\bibitem[{Chen et~al.(2020)Chen, Li, and Wang}]{chen-etal-2020-conditional}
Chen X, Li Q, Wang J (2020) Conditional causal relationships between emotions
  and causes in texts. In: Proceedings of the 2020 Conference on Empirical
  Methods in Natural Language Processing (EMNLP), Association for Computational
  Linguistics, Online, pp 3111--3121, \doi{10.18653/v1/2020.emnlp-main.252},
  \urlprefix\url{https://www.aclweb.org/anthology/2020.emnlp-main.252}

\bibitem[{Chen et~al.(2010)Chen, Lee, Li, and Huang}]{chen-etal-2010-emotion}
Chen Y, Lee SYM, Li S, Huang CR (2010) Emotion cause detection with linguistic
  constructions. In: Proceedings of the 23rd International Conference on
  Computational Linguistics (Coling 2010), Coling 2010 Organizing Committee,
  Beijing, China, pp 179--187,
  \urlprefix\url{https://www.aclweb.org/anthology/C10-1021}

\bibitem[{Chen et~al.(2018)Chen, Hou, Cheng, and
  Li}]{DBLP:conf/emnlp/ChenHCL18}
Chen Y, Hou W, Cheng X, Li S (2018) Joint learning for emotion classification
  and emotion cause detection. In: Riloff E, Chiang D, Hockenmaier J, Tsujii J
  (eds) Proceedings of the 2018 Conference on Empirical Methods in Natural
  Language Processing, Brussels, Belgium, October 31 -- November 4, 2018,
  Association for Computational Linguistics, pp 646--651,
  \doi{10.18653/v1/d18-1066},
  \urlprefix\url{https://doi.org/10.18653/v1/d18-1066}

\bibitem[{Choi et~al.(2005)Choi, Cardie, Riloff, and
  Patwardhan}]{DBLP:conf/naacl/ChoiCRP05}
Choi Y, Cardie C, Riloff E, Patwardhan S (2005) Identifying sources of opinions
  with conditional random fields and extraction patterns. In: {HLT/EMNLP} 2005,
  Human Language Technology Conference and Conference on Empirical Methods in
  Natural Language Processing, Proceedings of the Conference, 6-8 October 2005,
  Vancouver, British Columbia, Canada, The Association for Computational
  Linguistics, pp 355--362,
  \urlprefix\url{https://www.aclweb.org/anthology/H05-1045/}

\bibitem[{Colneri{\^c} and Demsar(2018)}]{colneric2018emotion}
Colneri{\^c} N, Demsar J (2018) Emotion recognition on twitter: comparative
  study and training a unison model. IEEE Transactions on Affective Computing

\bibitem[{Das and Bandyopadhyay(2010)}]{das-bandyopadhyay-2010-finding}
Das D, Bandyopadhyay S (2010) Finding emotion holder from {B}engali blog
  {T}exts{---}{A}n unsupervised syntactic approach. In: Proceedings of the 24th
  Pacific Asia Conference on Language, Information and Computation, Institute
  of Digital Enhancement of Cognitive Processing, Waseda University, Tohoku
  University, Sendai, Japan, pp 621--628,
  \urlprefix\url{https://www.aclweb.org/anthology/Y10-1071}

\bibitem[{Devlin et~al.(2019)Devlin, Chang, Lee, and
  Toutanova}]{devlin2018bert}
Devlin J, Chang M, Lee K, Toutanova K (2019) {BERT}: Pre-training of deep
  bidirectional transformers for language understanding. In: Burstein J, Doran
  C, Solorio T (eds) Proceedings of the 2019 Conference of the North American
  Chapter of the Association for Computational Linguistics: Human Language
  Technologies, {NAACL-HLT} 2019, Minneapolis, MN, USA, June 2-7, 2019, Volume
  1 (Long and Short Papers), Association for Computational Linguistics, pp
  4171--4186, \doi{10.18653/v1/n19-1423},
  \urlprefix\url{https://doi.org/10.18653/v1/n19-1423}

\bibitem[{Ding et~al.(2020{\natexlab{a}})Ding, Xia, and
  Yu}]{DBLP:conf/acl/DingXY20}
Ding Z, Xia R, Yu J (2020{\natexlab{a}}) {ECPE-2D}: Emotion-cause pair
  extraction based on joint two-dimensional representation, interaction and
  prediction. In: Jurafsky D, Chai J, Schluter N, Tetreault JR (eds)
  Proceedings of the 58th Annual Meeting of the Association for Computational
  Linguistics, {ACL} 2020, Online, July 5-10, 2020, Association for
  Computational Linguistics, pp 3161--3170,
  \doi{10.18653/v1/2020.acl-main.288},
  \urlprefix\url{https://doi.org/10.18653/v1/2020.acl-main.288}

\bibitem[{Ding et~al.(2020{\natexlab{b}})Ding, Xia, and
  Yu}]{DBLP:conf/emnlp/DingXY20}
Ding Z, Xia R, Yu J (2020{\natexlab{b}}) End-to-end emotion-cause pair
  extraction based on sliding window multi-label learning. In: Webber B, Cohn
  T, He Y, Liu Y (eds) Proceedings of the 2020 Conference on Empirical Methods
  in Natural Language Processing, {EMNLP} 2020, Online, November 16-20, 2020,
  Association for Computational Linguistics, pp 3574--3583,
  \doi{10.18653/v1/2020.emnlp-main.290},
  \urlprefix\url{https://doi.org/10.18653/v1/2020.emnlp-main.290}

\bibitem[{Dragoni et~al.(2021)Dragoni, Donadello, and Cambria}]{OntoSenticNet2}
Dragoni M, Donadello I, Cambria E (2021) {OntoSenticNet} 2: Enhancing reasoning
  within sentiment analysis. IEEE Intelligent Systems 36(5)

\bibitem[{Ekman(1993)}]{ekman1993facial}
Ekman P (1993) Facial expression and emotion. American Psychologist 48(4):384

\bibitem[{Ellsworth and Scherer(2003)}]{ellsworth2003appraisal}
Ellsworth PC, Scherer KR (2003) Appraisal processes in emotion, Oxford
  University Press, pp 572--595

\bibitem[{Gao et~al.(2017)Gao, Jiannan, Ruifeng, Lin, He, Wong, and
  Lu}]{gao2017overview}
Gao Q, Jiannan H, Ruifeng X, Lin G, He Y, Wong K, Lu Q (2017) Overview of
  ntcir-13 eca task. In: Proceedings of the NTCIR-13 Conference

\bibitem[{Ghazi et~al.(2015)Ghazi, Inkpen, and
  Szpakowicz}]{DBLP:conf/cicling/GhaziIS15}
Ghazi D, Inkpen D, Szpakowicz S (2015) Detecting emotion stimuli in
  emotion-bearing sentences. In: Gelbukh AF (ed) Computational Linguistics and
  Intelligent Text Processing -- 16th International Conference, CICLing 2015,
  Cairo, Egypt, April 14-20, 2015, Proceedings, Part {II}, Springer, Lecture
  Notes in Computer Science, vol 9042, pp 152--165,
  \doi{10.1007/978-3-319-18117-2\_12},
  \urlprefix\url{https://doi.org/10.1007/978-3-319-18117-2\_12}

\bibitem[{Ghosal et~al.(2020)Ghosal, Majumder, Mihalcea, and
  Poria}]{ghosal2020utterancelevel}
Ghosal D, Majumder N, Mihalcea R, Poria S (2020) Utterance-level dialogue
  understanding: An empirical study. \eprint{2009.13902}

\bibitem[{Gui et~al.(2014)Gui, Yuan, Xu, Liu, Lu, and
  Zhou}]{DBLP:conf/nlpcc/GuiYXLLZ14}
Gui L, Yuan L, Xu R, Liu B, Lu Q, Zhou Y (2014) Emotion cause detection with
  linguistic construction in {Chinese} {Weibo} text. In: Zong C, Nie J, Zhao D,
  Feng Y (eds) Natural Language Processing and Chinese Computing -- Third {CCF}
  Conference, {NLPCC} 2014, Shenzhen, China, December 5-9, 2014. Proceedings,
  Springer, Communications in Computer and Information Science, vol 496, pp
  457--464, \doi{10.1007/978-3-662-45924-9\_42},
  \urlprefix\url{https://doi.org/10.1007/978-3-662-45924-9\_42}

\bibitem[{Gui et~al.(2016)Gui, Wu, Xu, Lu, and Zhou}]{gui2016event}
Gui L, Wu D, Xu R, Lu Q, Zhou Y (2016) Event-driven emotion cause extraction
  with corpus construction. In: EMNLP, World Scientific, pp 1639--1649

\bibitem[{Izard(1992)}]{Izard1992}
Izard CE (1992) Basic emotions, relations among emotions, and emotion-cognition
  relations. Psychological Review 99(3):561--565,
  \doi{10.1037/0033-295X.99.3.561},
  \urlprefix\url{https://doi.org/10.1037/0033-295X.99.3.561}

\bibitem[{Joshi et~al.(2020)Joshi, Chen, Liu, Weld, Zettlemoyer, and
  Levy}]{joshi2020spanbert}
Joshi M, Chen D, Liu Y, Weld DS, Zettlemoyer L, Levy O (2020) {SpanBERT}:
  Improving pre-training by representing and predicting spans.
  \eprint{1907.10529}

\bibitem[{Kratzwald et~al.(2018)Kratzwald, Ilic, Kraus, Feuerriegel, and
  Prendinger}]{kratzwald2018decision}
Kratzwald B, Ilic S, Kraus M, Feuerriegel S, Prendinger H (2018) Decision
  support with text-based emotion recognition: Deep learning for affective
  computing. arXiv preprint arXiv:180306397

\bibitem[{Lee et~al.(2010)Lee, Chen, and Huang}]{lee-etal-2010-text}
Lee SYM, Chen Y, Huang CR (2010) A text-driven rule-based system for emotion
  cause detection. In: Proceedings of the {NAACL} {HLT} 2010 Workshop on
  Computational Approaches to Analysis and Generation of Emotion in Text,
  Association for Computational Linguistics, Los Angeles, CA, pp 45--53,
  \urlprefix\url{https://www.aclweb.org/anthology/W10-0206}

\bibitem[{Levy et~al.(2017)Levy, Seo, Choi, and
  Zettlemoyer}]{levy-etal-2017-zero}
Levy O, Seo M, Choi E, Zettlemoyer L (2017) Zero-shot relation extraction via
  reading comprehension. In: Proceedings of the 21st Conference on
  Computational Natural Language Learning ({C}o{NLL} 2017), Association for
  Computational Linguistics, Vancouver, Canada, pp 333--342,
  \doi{10.18653/v1/K17-1034},
  \urlprefix\url{https://www.aclweb.org/anthology/K17-1034}

\bibitem[{Li et~al.(2020)Li, Feng, Meng, Han, Wu, and
  Li}]{li-etal-2020-unified}
Li X, Feng J, Meng Y, Han Q, Wu F, Li J (2020) A unified {MRC} framework for
  named entity recognition. In: Proceedings of the 58th Annual Meeting of the
  Association for Computational Linguistics, Association for Computational
  Linguistics, Online, pp 5849--5859, \doi{10.18653/v1/2020.acl-main.519},
  \urlprefix\url{https://www.aclweb.org/anthology/2020.acl-main.519}

\bibitem[{Li et~al.(2017)Li, Su, Shen, Li, Cao, and Niu}]{li2017DailyDialog}
Li Y, Su H, Shen X, Li W, Cao Z, Niu S (2017) Dailydialog: {A} manually
  labelled multi-turn dialogue dataset. In: Kondrak G, Watanabe T (eds)
  Proceedings of the Eighth International Joint Conference on Natural Language
  Processing, {IJCNLP} 2017, Taipei, Taiwan, November 27 - December 1, 2017 --
  Volume 1: Long Papers, Asian Federation of Natural Language Processing, pp
  986--995, \urlprefix\url{https://www.aclweb.org/anthology/I17-1099/}

\bibitem[{Liu(2012)}]{liu2012sentiment}
Liu B (2012) Sentiment Analysis and Opinion Mining. Synthesis Lectures on Human
  Language Technologies, Morgan {\&} Claypool Publishers,
  \doi{10.2200/S00416ED1V01Y201204HLT016},
  \urlprefix\url{https://doi.org/10.2200/S00416ED1V01Y201204HLT016}

\bibitem[{Liu et~al.(2019)Liu, Ott, Goyal, Du, Joshi, Chen, Levy, Lewis,
  Zettlemoyer, and Stoyanov}]{liu2019roberta}
Liu Y, Ott M, Goyal N, Du J, Joshi M, Chen D, Levy O, Lewis M, Zettlemoyer L,
  Stoyanov V (2019) {RoBERTa}: A robustly optimized {BERT} pretraining
  approach. arXiv preprint arXiv:190711692

\bibitem[{{Moreno Jim\'enez} and {Torres Moreno}(2020)}]{LiSSS}
{Moreno Jim\'enez} LG, {Torres Moreno} JM (2020) {LiSSS}: A new corpus of
  literary {Spanish} sentences for emotions detection. Computaci\'on y Sistemas
  24(3):1139--1147, \doi{10.13053/CyS-24-3-3474}

\bibitem[{Neviarouskaya and Aono(2013)}]{DBLP:conf/ijcnlp/NeviarouskayaA13}
Neviarouskaya A, Aono M (2013) Extracting causes of emotions from text. In:
  Sixth International Joint Conference on Natural Language Processing, {IJCNLP}
  2013, Nagoya, Japan, October 14-18, 2013, Asian Federation of Natural
  Language Processing / {ACL}, pp 932--936,
  \urlprefix\url{https://www.aclweb.org/anthology/I13-1121/}

\bibitem[{Plutchik(1982)}]{plutchik}
Plutchik R (1982) A psychoevolutionary theory of emotions. Social Science
  Information 21(4-5):529--553, \doi{10.1177/053901882021004003}

\bibitem[{Rajpurkar et~al.(2016)Rajpurkar, Zhang, Lopyrev, and
  Liang}]{rajpurkar2016squad}
Rajpurkar P, Zhang J, Lopyrev K, Liang P (2016) Squad: 100,000+ questions for
  machine comprehension of text. \eprint{1606.05250}

\bibitem[{Talmy(2000)}]{talmy2000toward}
Talmy L (2000) Toward a cognitive semantics, vol~2. MIT press

\bibitem[{Wei et~al.(2020)Wei, Zhao, and Mao}]{wei-etal-2020-effective}
Wei P, Zhao J, Mao W (2020) Effective inter-clause modeling for end-to-end
  emotion-cause pair extraction. In: Proceedings of the 58th Annual Meeting of
  the Association for Computational Linguistics, Association for Computational
  Linguistics, Online, pp 3171--3181, \doi{10.18653/v1/2020.acl-main.289},
  \urlprefix\url{https://www.aclweb.org/anthology/2020.acl-main.289}

\bibitem[{Xia and Ding(2019)}]{DBLP:conf/acl/XiaD19}
Xia R, Ding Z (2019) Emotion-cause pair extraction: {A} new task to emotion
  analysis in texts. In: Korhonen A, Traum DR, M{\`{a}}rquez L (eds)
  Proceedings of the 57th Conference of the Association for Computational
  Linguistics, {ACL} 2019, Florence, Italy, July 28 -- August 2, 2019, Volume
  1: Long Papers, Association for Computational Linguistics, pp 1003--1012,
  \doi{10.18653/v1/p19-1096},
  \urlprefix\url{https://doi.org/10.18653/v1/p19-1096}

\bibitem[{Zajonc(1980)}]{Zajonc80feelingand}
Zajonc RB (1980) Feeling and thinking: Preferences need no inferences. American
  Psychologist pp 151--175

\end{thebibliography}

\end{document}